\documentclass[lettersize,journal]{IEEEtran}
\usepackage{amsmath,amsfonts}
\usepackage{algorithmic}
\usepackage{algorithm}
\usepackage{array}
\usepackage[caption=false,font=normalsize,labelfont=sf,textfont=sf]{subfig}
\usepackage{textcomp}
\usepackage{stfloats}
\usepackage{url}
\usepackage{verbatim}
\usepackage{graphicx}
\usepackage{cite}

\usepackage{amsmath,amssymb,amsfonts}
\usepackage{algorithmic}
\usepackage{textcomp}
\usepackage{dsfont}
\usepackage{enumerate}
\usepackage{comment}
\usepackage[english]{babel}
\usepackage{ragged2e}
\usepackage{cuted}
\usepackage{flushend}
\usepackage[usenames,dvipsnames]{color}

\newtheorem{theorem}{\normalfont \textbf{Theorem}}
\newtheorem{lemma}{\normalfont \textbf{Lemma}}

\newtheorem{corollary}{\normalfont \textbf{Corollary}}

\usepackage{mathtools}

\begin{document}

\title{New Bounds on the Accuracy of Majority Voting for Multi-Class Classification}

\author{Sina Aeeneh, Nikola Zlatanov,~\IEEEmembership{Member,~IEEE,} Jiangshan Yu,~\IEEEmembership{Member,~IEEE} 
\thanks{Sina Aeeneh is with the Department of Electrical and Computer Systems Engineering, Monash University, Clayton, VIC 3800, Australia (e-mail: sina.aeeneh@monash.edu).}
\thanks{Nikola Zlatanov is with the Department of Computer Science and Engineering, Innopolis University, Innopolis, Respublika Tatarstan, Russia, 420500.}
\thanks{Jiangshan Yu is with the Department of Software Systems and Cybersecurity, Monash University, Clayton, VIC 3800, Australia.}
}

\maketitle

\begin{abstract}
Majority voting is a simple mathematical function that returns the value that appears most often in a set. As a popular decision fusion technique, the majority voting function (MVF) finds applications in resolving conflicts, where a number of independent voters report their opinions on a classification problem. Despite its importance and its various applications in ensemble learning, data crowd-sourcing, remote sensing, and data oracles for blockchains, the accuracy of the MVF for the general multi-class classification problem has remained unknown. In this paper, we derive a new upper bound on the accuracy of the MVF for the multi-class classification problem. More specifically, we show that under certain conditions, the error rate of the MVF exponentially decays toward zero as the number of independent voters increases. Conversely, the error rate of the MVF exponentially grows with the number of independent voters if these conditions are not met.

We first explore the problem for independent and identically distributed voters where we assume that every voter follows the same conditional probability distribution of voting for different classes, given the true classification of the data point. Next, we extend our results for the case where the voters are independent but non-identically distributed. Using the derived results, we then provide a discussion on the accuracy of the truth discovery algorithms. We show that in the best-case scenarios, truth discovery algorithms operate as an amplified MVF and thereby achieve a small error rate only when the MVF achieves a small error rate, and vice versa, achieve a large error rate when the MVF also achieves a large error rate. In the worst-case scenario, the truth discovery algorithms may achieve a higher error rate than the MVF. Finally, we confirm our theoretical results using numerical simulations. 
\end{abstract}

\begin{IEEEkeywords}
Majority voting, data fusion, multi-class classification, crowd-sourcing, ensemble learning, truth discovery.
\end{IEEEkeywords}

\section{Introduction}
\IEEEPARstart{M}{ajority voting} is a simple decision rule that is used to combine multiple opinions in a classification task. The majority voting function (MVF) can be implemented in different variations including unanimity majority voting, simple majority voting, weighted majority voting, and plurality majority voting \cite{BookEnsamble}. From these, plurality majority voting has been the most commonly used form of the MVF \cite{BookEnsamble} and is the focus of this paper. The plurality MVF, or the MVF for short in the rest of the paper, receives votes from several voters on a specific classification problem and returns the class that has received the highest number of votes.
In practice, humans use majority voting to resolve disagreements in collective decision-making processes such as elections. Computer systems also employ majority voting to improve their performance in statistical machine learning \cite{Bagging, AdaBoost, StackedGeneralization}, data crowd-sourcing \cite{Crowdsourcing}, remote sensing \cite{MultiSensor, RemoteSensing}, and, more recently, in off-chain data oracles for blockchain-based smart contracts \cite{ASTRAEA, ChainLink-whitepaper,576970}. The MVF and its variations belong to a larger set of techniques for combining different opinions known as Decision Fusion Techniques (DFTs) \cite{DecsionFusionSurveyPattern}. The voters participating in a decision fusion task together with a DFT form a decision fusion system.

Ensemble classifiers are an example of a decision fusion system. Ensemble classifiers consist of several weak classifiers and a DFT, such as the MVF, that is used to combine the outputs of the base classifiers. Ensemble classifiers are used as the state-of-the-art approach to achieve a more accurate and more generalized classification model \cite{sagi2018ensemble}. In their seminal work \cite{Boosting}, the author introduced a novel technique called boosting to enhance the performance of classifiers. Boosting involves training an ensemble of weak classifiers using distinct subsets of the training data set. The author in \cite{Boosting} showed that training a large number of weak classifiers coupled with a proper DFT to aggregate their outputs could result in an arbitrarily accurate classification model, under certain conditions. 
Later on, several authors proposed similar algorithms such as bagging, AdaBoost, and stacked generalization to improve the accuracy and the generalization of the system \cite{Bagging, AdaBoost, StackedGeneralization}. The difference between various ensemble classification algorithms lies in the way they sample the training data used for the training of the base classifiers and in the way they combine the outputs of the base classifiers. For instance, while \cite{Bagging} uses the MVF for combining the outputs of the base classifiers, in \cite{StackedGeneralization} another classifier is trained to combine the outputs of the base classifiers and obtain the final output of the system. 

The MVF, and the DFTs in general, also find applications in crowd-sourcing systems \cite{Crowdsourcing}, where a number of independent agents try to label samples from a data set. The labeled data set is often needed for training machine learning algorithms. 
While individual agents participating in a crowd-sourcing task could be unreliable and noisy, the combination of their opinions using a suitable DFT may result in a data set labeled with high-accuracy \cite{Crowdsourcing}. Similarly, multi-sensor systems also use a DFT to combine the observations of several independent and possibly erroneous sensors in order to obtain a reliable classification result. For instance, the authors in \cite{MultiSensor} study the performance of a network of body-worn acceleration sensors in recognizing human gestures. They showed that combining the classification results of many sensors using MVF significantly improves the accuracy of the system. In another study, the authors use DFTs to combine the outputs of different types of sensors in order to obtain a more accurate classification result in remote sensing applications \cite{RemoteSensing}.

More recently, the idea of employing DFTs for aggregating conflicting opinions has been studied in the blockchain literature \cite{ASTRAEA, ChainLink-whitepaper,576970}. 
Since blockchains are isolated systems, blockchain-specific data-feed systems have been developed to read off-chain data from sensors, web pages, and human users, and record them on the blockchains when needed. 
In order to increase the accuracy and reliability of their services, some data-feed systems utilize multiple oracles for each data query task \cite{ASTRAEA, ChainLink-whitepaper}. Since disagreement between oracles may occur, a DFT is implemented to aggregate the reported values and obtain a unique value for each off-chain data query. ASTRAEA and Chainlink are examples of blockchain-based data-feed systems that employ the MVF to determine the final value of off-chain data queries \cite{ASTRAEA, ChainLink-whitepaper}.

The complexity of designing a robust DFT arises from the fact that in many applications the decision node performing the DFT does not have access to reliable information about the level of precision of the individual voters. Unlike the MVF which focuses on simplicity and completely disregards any potential side-information, a class of DFTs, commonly known as truth discovery algorithms, try to recursively estimate the reliability of different voters using the votes they have submitted during the previous classification tasks \cite{deducingtruth, TruthDiscoveryCorrelated}.
The estimated reliability index determines the weight of each voter and is updated at the beginning of each data fusion round. Once the voters propose their values for the requested parameter, the DFT combines them in proportion to the reliability index of each voter such that the opinion of a voter that is estimated to be more reliable has a higher weight in determining the final result.

The truth discovery solutions are considered to be more accurate when compared to the MVF \cite{wu2021blockchain, deducingtruth}.
At first glance, this seems to be correct since the truth discovery solutions exploit extra information, i.e., the past votes, to guess the accuracy of each voter. 
However, guessing the reliability of the voters from their past votes without having access to the past ground truths, and then, combining those past votes to generate reliability indexes, creates a positive feedback loop in these systems. When compared to the MVF, the positive feedback loop of the truth discovery algorithms amplifies the accuracy in the best-case scenario only when the MVF achieves a small error rate. However, the same positive feedback loop of the truth discovery algorithm amplifies the error rate in scenarios where the MVF achieves a high error rate. In other words, in terms of their error rate, the truth discovery algorithms are simply a MVF with an amplification factor. 

In this paper, we will focus on analyzing the accuracy of the MVF as a DFT under some general assumptions. Despite its simplicity and its practicality in many applications, there is a limited theoretical study on the accuracy of the MVF in the literature, most of which assume a binary classification problem, where the voters are allowed to vote either for \(0\) or for~\(1\). For instance, the authors in \cite{BlockchainOracleAdler} explored DFT for crowd-sourcing data into blockchains for binary data queries. Their analysis showed that if individual voters have an accuracy rate that is higher than \(0.5\), then a majority voting-based decision maker converges to the correct value with probability that grows towards \(1\) as the number of voters increases.
This is in line with the results in \cite{576970,ruta2002theoretical,kuncheva2003limits} which have investigated the MVF and derived its error rate for the binary classification problem. 
All of these studies, however, lack the ability to analyze the MVF in the context of the non-binary classification problem. The multi-class classifiers are more general than the binary classifiers, making them useful for a wider range of applications. However, analyzing the accuracy of a multi-class classifier is much more difficult, since the number of votes in favor of different classes are mutually dependent random variables. To the best of our knowledge, there are only two research papers that have theoretically studied the accuracy of the MVF for the non-binary classification problem
 \cite{Matan96onvoting, li2014error}.
The author in \cite{Matan96onvoting} derives an upper bound and a lower bound on the error rate of the MVF for the non-binary classification problem. However, \cite{Matan96onvoting} obtains bounds only for the strong MVF, a.k.a. the simple MVF, where at least half of the voters are needed to report a class in order for the MVF to be able to decide for that class. 
The strong MVF does not produce a decision if less than half of the voters voted for the most voted class, an event that is very frequent in multi-class classifiers. Therefore, the more commonly used plurality MVF that can achieve a much lower error rate than the strong MVF \cite{BookEnsamble} is not covered in \cite{Matan96onvoting}.
The authors in \cite{li2014error} propose another upper bound on the error rate of the MVF for the multi-class classification problem that is applicable to the plurality MVF. However, the upper bound in \cite{li2014error} is derived using the union bound technique and is, therefore, a weak bound in general that diverges from the actual error rate as the number of voters increases.

The absence of a comprehensive theoretical analysis regarding the reliability of MVF in multi-class classification presents a notable challenge when attempting to deploy it within decision fusion systems for real-world applications. For instance, in decentralized data-feed systems or redundant sensor networks, it is difficult to place trust in the accuracy of the output of the MVF without a formal analysis. In this paper, we theoretically analyze the accuracy of the MVF for the multi-class classification problem by utilizing results from the \textit{balls into bins} problem \cite{mitzenmacher} and applying them to the majority voting problem. The findings presented in this paper will assist system designers in evaluating the potential for achieving a low error rate through the use of MVF and in establishing the minimum number of voters necessary to guarantee a specific upper limit on the error rate.
We can summarize our main contributions as follows:
\begin{itemize}
    \item We derive an upper bound on the error rate of the MVF as a DFT assuming independent and identically distributed (i.i.d.) voters. The derived upper bound is shown to be tight in the numerical examples (Theorem \ref{theorem:1} in Section~III). 
    \item We derive the necessary and sufficient conditions under which the error rate of the MVF decays towards zero when the number of i.i.d. voters increases (Theorem \ref{prop1} and its corollaries in Section~III) in which case the slope of the decay of the error rate is exponential with the number of voters (Theorem \ref{theorem:11} in Section~III). 
    \item We generalize the results in Theorem \ref{theorem:1}, Theorem \ref{prop1}, and Theorem \ref{theorem:11} to independent non-identically distributed (i.non-i.d.) voters in Section~IV. More specifically, we derive an upper bound on the error rate of the system for the general case of i.non-i.d. voters (Theorem \ref{theorem:4} in Section IV), which is shown to be tight in the numerical examples. Moreover, we show that the error rate of the MVF for i.non-i.d. voters decays towards zero as the number of voters increases if, and only if, certain conditions are met (Theorem \ref{prop:10} and its corollaries in Section IV). We also show that under those conditions the slope of the decay of the error rate is exponential with the number of voters (Theorem \ref{prop:3} in Section~IV). 
    \item Leveraging the results obtained on the accuracy of the MVF, we compare a large class of DFTs, known as truth discovery algorithms, with the MVF and provide insights~(Section~V).
    \item Finally we provide numerical analysis for our upper bounds on the error rate of the MVF (Section~VI) and conclude the paper (Section~VII).
\end{itemize}

\textit{Notation:} The set \(\{1, 2, \dots , N\}\) is denoted by \([N]\). The notation \([N]  \setminus m\) is used to denote \(\{n|\,n \in [N]  \text{ and }  n\neq m\}\). We use uppercase symbols, such as \(X\), to denote random variables and lowercase symbols, such as \(x\), for their realizations. Moreover, we use the calligraphic font to denote sets, e.g. \(\mathcal{X}=\{X_1, X_2,\dots, X_N\}\), and the bold font to denote vectors and matrices, e.g., \(\mathbf{X}=\left(X_1, X_2,\dots, X_N\right)\). We use \(\mathcal{E}\left( . \right)\) to denote the occurrence of probabilistic events. Given two random variables, \(A\) and \(B\), the probability of \(\mathcal{E}\left(A=a\right)\) and probability of \(\mathcal{E}\left(A=a\right)\) given \(\mathcal{E}\left(B=b\right)\), are denoted by \(\textrm{Pr}\left(A=a\right)\) and \(\textrm{Pr}\left(A=a|B=b\right)\), respectively. Moreover, we use \(\left(\bigcap_{k=1}^{K} A_k=a_k\right)\) and \(\left(\bigcup_{k=1}^{K} A_k=a_k\right)\) to respectively denote the intersection and the union of events \(\mathcal{E}\left(A_k=a_k\right)\), $\forall k \in [K]$. The notation \(|.|\) is used for denoting both the absolute value of variables and the number of elements in sets. Notations \(\lfloor{.}\rfloor\) and \(\lceil{.}\rceil\) denote the flooring and ceiling functions, respectively.
Moreover, we use \(\mathbf{1}\left(.\right)\) to denote the indicator function. Notation \(O\left(.\right)\) is used to denote the order of a function. Finally, the modified Bessel function of the first kind of order \(\alpha\) is denoted by \(I_{\left|\alpha \right|}\left(.\right)\), where 
\begin{equation*}
    I_{\left|\alpha\right|}\left(x\right) = \sum_{m=0}^{\infty} \frac{1}{m!\left(m+|\alpha|\right)!}\left(\frac{x}{2}\right)^{2m+|\alpha|},
\end{equation*}
for any real positive \(x\) and integer \(\alpha\).

\section{System Model}
Let \(X\) represent the true classification of a data point that the data fusion system is trying to classify. We model \(X\) as a random variable with probability distribution \(p\left(x\right)\).
Let \(M\) denote the number of voters in the data fusion system. We denote the \(m\)'th voter by \(O_m\) and its subjective opinion about the value of \(X\) by \(Y_m\). The deviation of \(Y_m\) from \(X\) is modeled by a conditional probability distribution \(p\left(y_m|x\right)\).

\begin{figure}[!htb]
    \centering
    \includegraphics[width=0.49\textwidth]{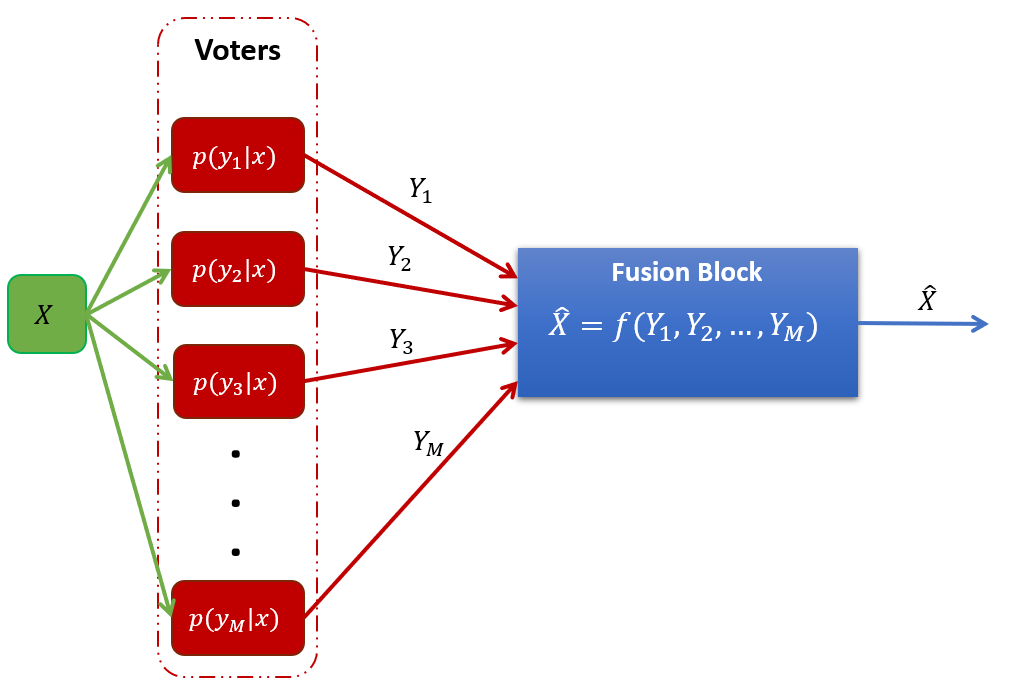}
   \caption{A sketch of the considered system model.}
   \label{fig:sketch}
\end{figure}

The conditional probability distribution captures the intrinsic inaccuracy of the \(m\)'th voter resulting from imperfect data reading or imperfect processing of data as well as the potential malicious behavior of the voter.
Next, each voter sends its subjective value \(Y_m, \forall m \in [M]\), to a data fusion block. Then, the data fusion block uses a function, \(f\left(.\right)\), defined as
\begin{equation}
    f:\left(Y_1, Y_2, \dots,Y_M\right)\to \hat{X},
\end{equation}
in order to obtain \(\hat{X}\), which is the estimation of the true parameter \(X\). Finally, the value of \(\hat{X}\) is reported as the final output of the data fusion system. A sketch of the considered system is shown in Fig. \ref{fig:sketch}. 

In the rest of this paper, we will focus on decision fusion systems that employ the MVF in the data fusion block, and thereby \(f\left(.\right)\) is given by 
\begin{equation}\label{eq:7} 
    f\left(Y_1, Y_2, \dots, Y_M\right) 
    =\underset{x_k}{\textrm{argmax}} \sum_{m=1}^{M} \mathbf{1}\left(Y_m = x_k\right).
\end{equation}
Hence, the MVF first counts the number of voters that have voted for \(x_k\), \(\forall k \in [K]\), and then, at the output reports the class that has received the highest number of votes\footnote{In the case of a tie between two or more classes, we break the tie by selecting one of the classes uniformly at random. Other rules are applicable.}.

Throughout this paper we assume that \(X\) takes values from the finite set \(\mathcal{X}=\left\{x_1,x_2, \dots, x_K\right\}\). Similarly, we assume that \(\hat{X}\), and \(Y_m\), \(\forall m \in [M]\), take values from the same set \(\mathcal{X}\). 

We use \(p_{l|k}^{\left(m\right)}\) to denote the probability that a voter \(O_m\) reports \(Y_m=x_l\) while \(X=x_k\), i.e., 
\begin{equation}
    p_{l|k}^{\left(m\right)}= \textrm{Pr}\left(Y_m=x_l|X=x_k \right).
\end{equation}

Our criteria for evaluating the performance of the data fusion system is the error rate defined as
\begin{equation}\label{eq:4}
    P_e= \textrm{Pr}\left(X\neq \hat{X}\right). 
\end{equation}
Hence, the optimal DFT is a DFT that minimizes \(P_e\).

For the worst case scenario in terms of minimising \(P_e\), we assume that \(X\) has a uniform distribution over its domain, i.e., \(\textrm{Pr}\left(X=x_k\right)=\frac{1}{K}\), \(\forall k \in \left[K\right]\). Moreover, we assume that the voters have independent observations of \(X\), i.e., 
\begin{multline}
    \textrm{~Pr}\left(Y_m=y_m,Y_n=y_n|X=x\right) =\textrm{Pr}\left(Y_m=y_m|X=x\right) \\ 
    \times \textrm{Pr}\left(Y_n=y_n|X=x\right),
    \text{~}\forall\, m, n \in [M].
\end{multline}    
This condition holds when the voters are not colluding and/or copying data from each other.

Generally, different voters may have different conditional probabilities due to the different data sources that they have access to as well as the different algorithms they may use to process data. However, in order to make the presentation easier to follow, we first investigate the case when the voters' conditional probabilities, \(p\left(y_m|x\right)\)'s, are i.i.d., and then we extend our results for independent but non-identically distributed (i.non-i.d.) voters.

\section{Upper Bounds On The Error Rate Of The Majority Voting Function For Independent And Identically Distributed Voters}
In this section, we derive an upper bound on the error rate of the MVF for i.i.d. voters. More precisely, we assume that 
\begin{itemize}
    \item the votes are independent, i.e., \( \textrm{Pr}\left(Y_m,Y_n|X\right)=\textrm{Pr}\left(Y_m|X\right)\textrm{Pr}\left(Y_n|X\right)\), 
    \item and the votes are identically distributed, i.e., \(p_{l|k}^{\left(m\right)}=p_{l|k}^{\left(n\right)}, \text{~}\forall m,n \in [M]\).
\end{itemize}
Since we are assuming that the voters are i.i.d, the value of \(p_{l|k}^{\left(m\right)}\) does not depend on the index \(m\). Therefore, we simplify the notation and denote the conditional probability of any voter voting for \(Y_m=x_l\), while \(X=x_k\), by \(p_{l|k}\), in the rest of this section.

\subsection{The Upper Bound On The Error Rate}
Our main result in this subsection is presented in Theorem~\ref{theorem:1}, which gives an upper bound on the error rate of a DFT employing the MVF for i.i.d. voters. To this end, we first define two sets of auxiliary variables and a lemma that we need in order to prove Theorem \ref{theorem:1}. 

Let \(\mathbf{V}=\left(V_1,V_2,\dots,V_K\right)\) be a random vector, where its \(k\)'th element, \(V_k\), denotes the number of votes submitted for class \(x_k\), and is given by 
\begin{equation*}
    V_k= \sum_{m=1}^{M} \mathbf{1} \left(Y_m=x_k\right).
\end{equation*}
\sloppypar{The conditional probability distribution of \(\mathbf{V}\) given that \(X=x_k\) is a Multinomial distribution given by} 
\begin{align}\label{eq:29}
    \textrm{Pr}\left(\mathbf{V}=\mathbf{v}|X=x_k\right) = \frac{M!}{v_1!\,v_2!\,\dots \,v_K!} \prod_{l=1}^{K} p_{l|k}^{v_l},
\end{align}
where $v_l$ denotes the realization of $V_l$ and $\sum_{l=1}^K v_l =M$.

As can be seen from \eqref{eq:29}, the random variables (RVs) \(V_1, V_2,\dots, V_K\) are mutually dependent. This makes the derivation of the error rate intractable. 
In order to tackle this problem, we define \(\mathbf{\hat{V}}=\left(\hat{V_1},\hat{V_2},\dots,\hat{V_K}\right)\) as a vector of RVs, where given \(X=x_k\), the RV \(\hat{V_l} \in \mathbf{\hat{V}}\) is independent of the rest of the elements in \(\mathbf{\hat{V}}\) and follows a Poisson distribution with parameter \(\lambda=M p_{l|k}\), given by
\begin{align}\label{eq:15}
    \textrm{Pr}\left(\hat{\mathbf{V}}=\hat{\mathbf{v}}|X=x_k\right) &= \prod_{l=1}^{K} \textrm{Pr}\left(\hat{V_l}=\hat{v_l}|X=x_k\right)\nonumber \\
        &=\prod_{l=1}^{K} \frac{e^{-M p_{l|k}} \left(M p_{l|k}\right)^{\hat{v_l}}}{\hat{v_l} !}.
\end{align}

We now give an important lemma that connects the probability events of \(\mathbf{V} \) to the probability events of \(\hat{\mathbf{V} }\). 
\begin{lemma} \label{lemma:4} 
Let \(\mathcal{E}\left(\mathbf{V} \right)\) be an event of a specific outcome of \(\mathbf{V}\) whose probability is either monotonically increasing or monotonically decreasing with the number of voters, \(M\). Similarly, let \(\mathcal{E}\left(\hat{\mathbf{V}} \right)\) denote the same event applied on the random vector \(\hat{\mathbf{V}}\). Then, the following holds
\begin{equation}
    \textrm{Pr}\left(\mathcal{E}\left(\mathbf{V} \right)\right) \leq 2 \textrm{Pr}\left(\mathcal{E}\left(\hat{\mathbf{V}} \right)\right).
\end{equation}
\end{lemma} 
\begin{IEEEproof} 
    Please refer to Appendix A. 
\end{IEEEproof}

Lemma \ref{lemma:4} plays an important role in our theoretical analysis throughout this paper by making it possible to substitute the dependent Multinomial RVs in \eqref{eq:29} with independent Poisson RVs in \eqref{eq:15} and obtain upper bounds on specific events. Following, we utilize this technique to derive an upper bound on the error rate of the MVF for i.i.d. voters.

\begin{theorem} \label{theorem:1}
The error rate of the majority voting DFT for i.i.d. voters is upper bounded by
\begin{align}\label{eq:upper-bound}
    P_e \leq 2 \Biggl(1-\frac{1}{K} \sum _{k=1} ^K \prod_{\substack{l=1 \\ l\neq k}}^K \Biggl(1-\sum_{\alpha=0}^\infty &e^{-M\left(p_{k|k}+p_{l|k}\right)}\left(\frac{p_{l|k}}{p_{k|k}}\right)^{\frac{\alpha}{2}} \nonumber\\    
    &\times I_{\left|\alpha\right|}\left(2M\sqrt{p_{l|k}p_{k|k}}\right)\Biggl) \Biggl),    
\end{align}
where \(I_{\left|\alpha\right|}\left(.\right)\) denotes the modified Bessel function of the first kind of order \(\left|\alpha\right|\).
\end{theorem}

\begin{IEEEproof}
The error rate of the MVF is given by
\begin{align}\label{eq:31}
    P_e                     &= \textrm{Pr}\left( \hat{X} \neq X\right)\nonumber\\
                            &= \frac{1}{K} \sum_{k=1}^K  \textrm{Pr}\left(\hat{X} \neq X|X=x_k\right)\nonumber\\
                            &\stackrel{(a)}{\leq} \frac{1}{K} \sum_{k=1}^K  \textrm{Pr}\left(\underset{l\neq k}{\bigcup_{l=1}^K} V_l \geq V_k \bigg| X=x_k\right) \nonumber\\
                            &\stackrel{(b)}{\leq} \frac{1}{K} \sum_{k=1}^K  2 \textrm{Pr}\left(\underset{l\neq k}{\bigcup_{l=1}^K} \hat{V}_l \geq
                            \hat{V}_k \bigg| X=x_k\right) \nonumber\\
                            &\stackrel{(c)}\leq 2\left(1- \frac{1}{K} \sum_{k=1}^K  \textrm{Pr}\left(\underset{ l\neq k}{\bigcap_{l=1} ^K} \hat{V}_l < \hat{V}_k \bigg| X=x_k\right)\right) \nonumber\\
                            &\stackrel{(d)}{=} 2\left(1- \frac{1}{K} \sum_{k=1}^K  \underset{l\neq k}{\prod_{l=1}^K} \textrm{Pr}\left( \hat{V}_l < \hat{V}_k \bigg| X=x_k\right)\right) \nonumber\\
                            &= 2\left(1- \frac{1}{K} \resizebox{.7 \hsize}{!}{\(\sum_{k=1}^K  \underset{l\neq k}{\prod_{l=1}^K} \left(1- \sum_{\alpha=0}^{\infty}  \textrm{Pr}\left( \hat{V}_l - \hat{V}_k =\alpha \bigg| X=x_k\right)\right)\)}\right) \nonumber\\
                            &\stackrel{(e)}{=} 2 \Biggl(1-\frac{1}{K} \sum _{k=1} ^K \prod_{\substack{l=1 \\ l\neq k}}^K \Biggl(1-\sum_{\alpha=0}^\infty e^{-M\left(p_{k|k}+p_{l|k}\right)}\left(\frac{p_{l|k}}{p_{k|k}}\right)^{\frac{\alpha}{2}} \nonumber\\
                            &\text{~~~~~~~~~~~~~~~~~~~~~~~~~~~~~~~~~~~~~}\times I_{\left|\alpha\right|}\left(2M\sqrt{p_{l|k}p_{k|k}}\right)\Biggl)\Biggl),
\end{align}
where the right-hand side of \((a)\) denotes the error rate assuming that MVF always breaks ties in favor of an incorrect class. Of course, this is a worst-case assumption. In practice, we can simply implement the MVF to break ties at random and achieve a lower error rate. However, in \((a)\) we bound the error rate of the MVF with the worst-case assumption for simplicity. Next, \((b)\) is a result of Lemma \ref{lemma:4}. Please note that the conditional error rate is a monotonic function of \(M\) since the voters are i.i.d. Hence, we can use Lemma \ref{lemma:4} to bound the error rate as per \((b)\). Next, \((c)\) is a result of the complement rule for probability, and \((d)\) is derived from the conditional independence of the elements in \(\hat{\mathbf{V}}\). Finally, \((e)\) is obtained by substituting \(\textrm{Pr}\left( \hat{V_l} - \hat{V_k} = \alpha | X=x_k\right)\) with the Skellam probability mass function, which is the probability mass function of the difference of two independent Poisson random variables \cite{Skellam}.
\end{IEEEproof}

The upper bound obtained in Theorem \ref{theorem:1} is in its most general form and may not provide intuition about its behavior. To provide intuition, we further simplify \eqref{eq:upper-bound} to explore the behavior of the upper bound in Theorem \ref{theorem:1} in both the asymptotic and transient cases.

\subsection{Asymptotic Behaviour Of The Error Rate}
In the following, we first state a theorem and then exploit it to simplify the upper bound in \eqref{eq:upper-bound} in the asymptotic scenario.
\begin{theorem}\label{prop1}
    \sloppy Assuming that \(p_{l_1|k} > p_{l_2|k}\) for given \(k, l_1,\text{~and~} l_2 \in [K]\), we can bound \(\textrm{Pr}\left(V_{l_1} \leq V_{l_2}|X=x_k\right)\) as
    \begin{equation}\label{eq:96}
        \textrm{Pr}\left(V_{l_1} \leq V_{l_2}|X=x_k\right)\leq 2e^{-M\left(\sqrt{p_{l_1|k}}-\sqrt{p_{l_2|k}} \right)^2}. 
    \end{equation}
\end{theorem}
\begin{IEEEproof}
    Please refer to Appendix B.
\end{IEEEproof}

\begin{corollary}\label{cor:1}
In the asymptotic regime, where \(M\to \infty\), the error rate, conditioned on \(X=x_k\), goes towards \(0\) if \(p_{k|k}>p_{l|k}\), \(\forall l\neq k\) and it goes towards \(1\) if \(\exists\, l_0 \neq k\) such that \(p_{k|k}<p_{l_0|k}\). 
\end{corollary}

\begin{IEEEproof}
Assuming that \(p_{l_1|k}>p_{l_2|k}\) for given \(k, l_1,\text{~and~} l_2 \in [K]\), we take the limit of both sides of the inequality in \eqref{eq:96} as
\begin{align}\label{eq:1001}
    \lim_{M\to \infty}\textrm{Pr}\left(V_{l_1} \leq V_{l_2}|X=x_k\right) &\leq \lim_{M\to \infty}2e^{-M\left(\sqrt{p_{l_1|k}}-\sqrt{p_{l_2|k}} \right)^2} \nonumber\\
                                                                         &= 0.
\end{align}
Using \eqref{eq:1001}, for the case where \(p_{k|k}>p_{l|k}\), \(\forall l\neq k\), we bound the error rate, conditioned on \(X=x_k\), as 
\begin{align}\label{eq:1201}
    \lim_{M\to \infty} P_{e|x_k} &\stackrel{(a)} {\leq} \lim_{M\to \infty}\textrm{Pr}\left(\underset{l\neq k}{\bigcup_{l=1}^K} V_l \geq V_k|X=x_k\right) \nonumber\\ 
    &\stackrel{(b)} {\leq} \lim_{M\to \infty} \underset{l\neq k}{\sum_{l=1}^K} \textrm{Pr}\left(V_{l}\geq V_{k}|X=x_k\right)\nonumber\\
    &\stackrel{(c)}{=} 0,
\end{align}
where the right-hand side of \((a)\) denotes the error rate assuming that MVF always breaks ties in favor of an incorrect class, which is a worst-case assumption. 
Moreover, \((b)\) is obtained using union bound, and \((c)\) is due to \eqref{eq:1001}.
Conversely, if \(\exists \, l_0 \neq k\) such that \(p_{k|k}<p_{l_0|k}\), we have
\begin{align}\label{eq:1202}
    \lim_{M\to \infty} P_{e|x_k} &\stackrel{(d)}{\geq} \lim_{M\to \infty}\textrm{Pr}\left(\underset{l\neq k}{\bigcup_{l=1}^K} V_l > V_k|X=x_k\right) \nonumber\\ 
    &\stackrel{(e)}{\geq} \lim_{M\to \infty} \textrm{Pr}\left(V_{l_0}> V_{k}|X=x_k\right) \nonumber\\
    &= 1- \lim_{M\to \infty} \textrm{Pr}\left(V_{l_0} \leq V_{k}|X=x_k\right)\nonumber\\
    &\stackrel{(f)}{=}1,
\end{align}
where the right-hand side of \((d)\) denotes the error rate assuming that the MVF always breaks ties in favor of the correct class \(x_k\). This is a best-case assumption. Therefore, the right-hand side of \((d)\) is a lower bound on the actual error rate. Moreover, \((e)\) results from the fact that \(\mathcal{E}\left(V_{l_0} > V_{k}\right)\) implies \(\mathcal{E}\left(\underset{l\neq k}{\bigcup_{l=1}^K} V_l > V_k \right)\).
Finally, \((f)\) is a result of \eqref{eq:1001} and the fact that \(p_{k|k}<p_{l_0|k}\). Combining \eqref{eq:1201} and \eqref{eq:1202} into one equation, we can describe the asymptotic behavior of the error rate, conditioned on \(X=x_k\), as 
\begin{align}\label{eq:27}
    \lim_{M\to \infty}P_{e|x_k} &\triangleq \lim_{M\to \infty} \textrm{Pr}\left(\hat{X}\neq x_k|X=x_k\right)\nonumber\\
                                &=\begin{cases}
                        0  & \text{if~} \forall \text{~} l\neq k, \text{~}p_{k|k}>p_{l|k}\\
                        1  & \text{if~} \exists \text{~} l\neq k, \text{~}p_{k|k}<p_{l|k}.\\
    \end{cases}
\end{align}
\end{IEEEproof}

\begin{corollary}\label{cor:2}
    Let \(\mathcal{A}\) be the largest subset of \([K]\) such that \(\forall k \in \mathcal{A}\) we have \(p_{k|k}>p_{l|k}\), \(\forall l\in[K]\). Then, the error rate of the MVF converges towards \(1-\frac{|\mathcal{A}|}{K}\) as \(M\) grows.
\end{corollary}
\begin{IEEEproof}
    The proof is straightforward. From \eqref{eq:27}, we have
\begin{align}\label{eq:11}
    \lim_{M\to \infty}P_e &=\lim_{M\to \infty}\sum_{k=1}^K \textrm{Pr}\left(X=x_k\right) P_{e|x_k} \nonumber\\
                          &= \sum_{k\in\mathcal{A}}\frac{1}{K} \times 0+ \sum_{k\in [K] \setminus \mathcal{A}} \frac{1}{K}  \nonumber\\
                          &=1-\frac{|\mathcal{A}|}{K}.     
\end{align}
\end{IEEEproof}
Corollary \ref{cor:2} is a generalization of the well-known Condorcet Jury Theorem \cite{BookEnsamble} that is originally stated for the binary MVF. Due to Corollary \ref{cor:2}, we need \(p_{k|k}>p_{l|k}\), \(\forall l,k \in [K]\), for a decision fusion system to work reliably, i.e.,
\begin{align}\label{eq:501}
    \lim_{M\to \infty}P_{e} = 0  \iff \forall l,k \in [K],\text{~where~} l\neq k, \text{~}p_{k|k}>p_{l|k}.
\end{align}

\subsection{Transient Behaviour Of The Error Rate}
In the following, we present a theorem that exploits the inequality in \eqref{eq:96} to simplify the upper bound obtained in Theorem~\ref{theorem:1} for the case when \(p_{k|k}>p_{l|k}\), \(\forall k,l \in [K]\). To this end, we first define two auxiliary variables, \(k^*\) and \(l^*\), as
\begin{align}\label{eq:926}
    \left(k^*,l^*\right)\triangleq \underset{k\neq l}{\underset{k,l}{\text{argmin}}}\left(\sqrt{p_{k|k}}- \sqrt{p_{l|k}} \right).
\end{align}
In words, \(\left(k^*,l^*\right)\) is the pair \(\left(k,l\right)\) that minimizes the difference \(\left(\sqrt{p_{k|k}}- \sqrt{p_{l|k}}\right)\), \(\forall \left(k,l\right) \in [K]\), where \(k\neq l\).

\begin{theorem}\label{theorem:11}
    If \(p_{k|k}>p_{l|k}, \forall k,l \in [K]\), then the upper bound in Theorem \ref{theorem:1} is simplified to
    \begin{equation}\label{eq:296}
        {P_e}\leq e^{-M\left(\left(\sqrt{p_{k^*|k^*}}-\sqrt{p_{l^*|k^*}} \right)^2 +\epsilon\right)}, 
    \end{equation}
where \(O(\epsilon)=\frac{\ln\left(M^{-3/4}\right)}{M}\).
\end{theorem}

\begin{IEEEproof}
    Please refer to Appendix C.
\end{IEEEproof}
As detailed in Appendix C, we derive the upper bound in Theorem \ref{theorem:11} by using $\textrm{Pr}\left(\hat{X}=x_{l^*} | X= x_{k^*}\right)$ to bound $P_e=\textrm{Pr}\left(\hat{X}\neq X\right)$. This is feasible because \(\left(k^*,l^*\right)\) corresponds to the pair of $\left(k,l\right)$ that maximizes $\textrm{Pr}\left(\hat{X}=x_{l} | X= x_{k}\right)$.

By taking the logarithm of both sides of \eqref{eq:296} and dividing the result by \(M\), we obtain
\begin{equation}\label{eq:901}
    \frac{\ln{P_e}}{M} \leq -\left(\sqrt{p_{k^*|k^*}}-\sqrt{p_{l^*|k^*}} \right)^2 +  \epsilon. 
\end{equation}
Therefore, assuming that the voters satisfy \(p_{k|k}>p_{l|k}, \forall k,l \in [K]\), \eqref{eq:901} shows that the slope of the decay is at most \(-\left(\sqrt{p_{k^*|k^*}}-\sqrt{p_{l^*|k^*}} \right)^2\).

As we have shown above, when \(p_{k|k}>p_{l|k}\), \(\forall l,k \in [K]\), the MVF achieves a very small error rate assuming that the number of participating voters, \(M\), is large enough. 
On the contrary, we observed that if for a given \(k\in [K]\), \(\exists l\neq k \in [K]\) such that \(p_{k|k}<p_{l|k}\) then the MVF almost surely makes an error in detecting \(X\) when \(X=x_k\).

\section{Upper Bounds On The Error Rate Of The Majority Voting Function For The Case Of Independent and Non-Identically Distributed Voters}
In this section, we extend the analysis in Section III to obtain upper bounds on the error rate of the MVF for the scenario when the voters are i.non-i.d. More precisely, throughout this section, we only assume that 
\begin{itemize}[]
    \item the voters are independent, i.e., \(\textrm{Pr}\left(Y_m,Y_n|X\right)=\textrm{Pr}\left(Y_m|X\right)\textrm{Pr}\left(Y_n|X\right)\).
\end{itemize}

Recalling from Section II that \(p_{l|k}^{\left(m\right)}\) denotes the probability that voter \(O_m\) reports \(Y_m=x_l\) while \(X=x_k\), we define a matrix, which we refer to it as the probability transition matrix for voter \(O_m\) and denote it by \(\textrm{p}^{(m)}\), as
\begin{align}\label{eq:23}
    &\textrm{p}^{\left(m\right)}=\begin{bmatrix} 
                    p_{1|1}^{\left(m\right)} & p_{1|2}^{\left(m\right)} &\dots &p_{1|K}^{\left(m\right)} \\
                    p_{2|1}^{\left(m\right)} & p_{2|2}^{\left(m\right)} &\dots &p_{2|K}^{\left(m\right)} \\
                     \vdots &\vdots  &\ddots    &\vdots \\
                    p_{K|1}^{\left(m\right)} &p_{K|2}^{\left(m\right)}   &\dots   &p_{K|K}^{\left(m\right)} 
                \end{bmatrix}.
\end{align}

\sloppy We partition the voters based on their probability transition matrices. To this end, let \(\left\{{ \mathcal{G}}_1, \mathcal{G}_2, \dots, \mathcal{G}_T\right\}\) be a partitioning on the set of voters, \(\mathcal{O}=\left\{O_1,O_2,\dots, O_M\right\}\), such that every element in \(\mathcal{O}\) falls in exactly one set \(\mathcal{G}_t\) and for any two voters, \(O_m\) and \(O_n\) that fall in the same set, we have \(\textrm{p}^{(m)}=\textrm{p}^{(n)}\). For simplicity of notation, the probability transition matrix of \(O_m, \forall \, O_m \in \mathcal{G}_t\) is denoted by \(\mathbf{q}^{(t)}\), where the \((k,l)\) element of \(\mathbf{q}^{(t)}\) is denoted by \(q_{l|k}^{\left(t\right)}\) and is given by
\begin{equation}
    q_{l|k}^{\left(t\right)}= p_{l|k}^{\left(m\right)},\, \forall\, m\in \mathcal{G}_t.
\end{equation}
The intuition behind partitioning voters into mutually disjoint sets is to capture the differences and similarities in their probability transition matrices. Here, we should emphasize that the MVF does not need to know the probability transition matrix of different voters; we use the notion of the probability transition matrix, and the partitioning introduced above, only for obtaining the upper bound on the error rate of the MVF.

We define random vector \(\mathbf{V}^{(t)}\) as
\begin{align*}   
    &\mathbf{V}^{(t)} \triangleq \left(V_1^{(t)}, V_2^{(t)}, \dots, V_K^{(t)}\right), \nonumber
\end{align*}
where, for each $k \in [K]$, \(V_k^{(t)}\) denotes the number of voters in \(\mathcal{G}_t\) that have voted for class \(x_k\).

Let \(r_t\) represent the proportion of voters in $\mathcal{G}_t$, i.e.,
\begin{equation*}
    r_t=\frac{|\mathcal{G}_t|}{M}.
\end{equation*}
The conditional probability distribution of \(\mathbf{V}^{(t)}=\left(V_1^{(t)}, V_2^{(t)}, \dots, V_K^{(t)}\right)\) given \(X=x_k\) is a Multinomial distribution, given by 
\begin{align}\label{eq:2011}
    \textrm{Pr}\left(\mathbf{V}^{(t)}=\mathbf{v}^{(t)} \Big | X=x_k \right) = \frac{(r_t M)!}{ \prod_{l=1}^K v_l^{(t)}!} \prod_{l=1}^K \left(q_{l|k}^{(t)}\right)^{v_l^{(t)}},
\end{align}
where \(\sum_{l=1}^K v_l^{(t)} =r_t M\).

The output of the MVF is determined by finding the maximum element of \(\mathbf{V}=\left(V_1, V_2, \dots, V_K\right)\) where its \(k\)'th element, \(V_k\), is given by
\begin{equation*}
    V_k \triangleq \sum_{t=1}^T V_k^{(t)}.
\end{equation*}
Please note that unlike what we had in Section III for i.i.d. voters, here the random vector \(\mathbf{V}\) does not necessarily follow a Multinomial distribution, only \(\mathbf{V}^{(t)}\) follows a Multinomial distribution, see \eqref{eq:2011}.

\subsection{The Upper Bound On The  Error Rate}
We state a lemma that is used in the proof of the upper bound on the error rate of the MVF for i.non-i.d. voters. First, we define an auxiliary variable, \(\delta_{l_1,l_2|k}\), as
\begin{equation}\label{eq:301}
    \delta_{l_1,l_2|k} \triangleq \sum_{t=1}^T r_t \left( q_{l_1|k}^{(t)}- q_{l_2|k}^{(t)}\right).
\end{equation} 
In words, \(\delta_{l_1,l_2|k}\) measures the difference between the likeliness of a randomly selected voter voting for class \(x_{l_1}\) and the likeliness that it votes for the class \(x_{l_2}\), given that \(X=x_k\). We use \(\delta_{l_2|k}\) instead of \(\delta_{k,l_2|k}\) for the special case where \(l_1=k\) in order to simplify the notation.

\begin{lemma}\label{lemma:5}
If \(\hat{X}\neq X\) when \(X= x_k\), then there exists at least one \(l\in [K]\) and one \(t \in [T]\) such that 
    \begin{equation}\label{eq:306}
         {v}_l^{(t)}-{v}_k^{(t)} \geq E\left\{{V}_l^{(t)}-{V}_k^{(t)}\right\}+r_t M \delta_{l|k},
    \end{equation}  
where \({v}_l^{(t)}\) and \({v}_k^{(t)}\) are the realizations of \({V}_l^{(t)}\) and \({V}_k^{(t)}\), respectively.
\end{lemma}
\begin{IEEEproof}
    Please refer to Appendix D.
\end{IEEEproof}
Lemma \ref{lemma:5} shows that for the MVF to make an incorrect decision when \(X=x_k\), the inequality in \eqref{eq:306} must hold for at least one \(t \in [T]\) and one \(l\in [K]\). Next, we exploit this lemma to obtain an upper bound on the error rate of the MVF for i.non-i.d. voters.

\begin{theorem}\label{theorem:4}
The error rate of the MVF for the case of i.non-i.d. voters is upper bounded by
\begin{align}\label{eq:76}
           P_e\leq \frac{2}{K} \sum_{k=1}^K \Biggl(1- {\prod_{t=1}^T}\underset{l\neq k}{\prod_{l=1}^K} \Biggl(1- \sum_{\alpha=\alpha_0}^\infty & \resizebox{.42 \hsize}{!}{\(e^{-r_t M \left(q_{k|k}^{\left(t\right)}+q_{l|k}^{\left(t\right)}\right)}\left(\frac{q_{l|k}^{\left(t\right)}}{q_{k|k}^{\left(t\right)}}\right)^{\frac{\alpha}{2}}\)} \nonumber\\
           & \times \resizebox{.3 \hsize}{!}{\(I_{\left|\alpha\right|}\left(2 r_t M\sqrt{q_{l|k}^{\left(t\right)}q_{k|k}^{\left(t\right)}}\right)\)} \Biggl) \Biggl),
\end{align}
where \(\alpha_0= r_t M \left({q_{l|k}^{(t)}}-{q_{k|k}^{(t)}} +\delta_{l|k}\right)\).
\end{theorem}

\begin{IEEEproof}
According to Lemma \ref{lemma:5}, \eqref{eq:306} needs to hold for at least one \(l\in[K]\) and one \(t\in[T]\) in order for the MVF to output an incorrect class. 
Starting from the definition of the error rate, \(P_e\), we use inequality \eqref{eq:306} to upper bound \(P_e\) as presented in \eqref{eq:307},
\begin{figure*}
  \hrulefill
\begin{align}\label{eq:307}
    P_e                     &{=} \textrm{Pr}\left(\hat{X} \neq X\right)\nonumber\\
                            &{=} \frac{1}{K} \sum_{k=1}^K  \textrm{Pr}\left(\hat{X} \neq X|X=x_k\right)\nonumber\\
                            &{\stackrel{(a)}{\leq}} \frac{1}{K} \sum_{k=1}^K  \textrm{Pr}\left(\underset{l\neq k}{\bigcup_{l=1}^K} V_l \geq V_k \Big|X=x_k\right)\nonumber\\
                            & {\stackrel{(b)}{=} \frac{1}{K} \sum_{k=1}^K \textrm{Pr} \left({\bigcup_{t=1}^T}\underset{l\neq k}{\bigcup_{l=1}^K} \left\{V_l^{(t)}-V_k^{(t)} \geq E\left\{V_l^{(t)}-V_k^{(t)}\right\}+r_t M \delta_{l|k}\right\} \Big|X=x_k\right)}       \nonumber\\              
                            & {\stackrel{(c)}{\leq} \frac{2}{K} \sum_{k=1}^K \textrm{Pr}\left({\bigcup_{t=1}^T}\underset{l\neq k}{\bigcup_{l=1}^K} \left\{\hat{V}_l^{(t)}-\hat{V}_k^{(t)} \geq E\left\{V_l^{(t)}-V_k^{(t)}\right\}+r_t M \delta_{l|k}\right\} \Big|X=x_k \right) }\nonumber\\ 
                            &\stackrel{}{=} \frac{2}{K} \sum_{k=1}^K \left(1- \textrm{Pr} \left({\bigcap_{t=1}^T}\underset{l\neq k}{\bigcap_{l=1}^K} \left\{\hat{V}_l^{(t)}-\hat{V}_k^{(t)} <
                            E\left\{V_l^{(t)}-V_k^{(t)}\right\}+r_t M \delta_{l|k}\right\} \Big|X=x_k \right)\right)        \nonumber\\      
                            &\stackrel{(d)}{=} \frac{2}{K} \sum_{k=1}^K \left(1- {\prod_{t=1}^T}\underset{l\neq k}{\prod_{l=1}^K} \textrm{Pr}\left(\hat{V}_l^{(t)}-\hat{V}_k^{(t)} <
                            E\left\{V_l^{(t)}-V_k^{(t)}\right\}+r_t M \delta_{l|k} \Big|X=x_k \right)\right)\nonumber\\  
                            &\stackrel{}{=} \frac{2}{K} \sum_{k=1}^K \left(1- {\prod_{t=1}^T}\underset{l\neq k}{\prod_{l=1}^K} \left(1- \sum_{\alpha=\alpha_0}^\infty \textrm{Pr}\left(\hat{V}_l^{(t)}-\hat{V}_k^{(t)} = \alpha \Big|X=x_k \right) \right)\right)  \nonumber\\ 
                            & \stackrel{(e)}{\leq} \frac{2}{K} \sum_{k=1}^K \left(1- {\prod_{t=1}^T}\underset{l\neq k}{\prod_{l=1}^K} \left(1- \sum_{\alpha=\alpha_0}^\infty e^{-r_t M \left(q_{k|k}^{\left(t\right)}+q_{l|k}^{\left(t\right)}\right)}\left(\frac{q_{l|k}^{\left(t\right)}}{q_{k|k}^{\left(t\right)}}\right)^{\frac{\alpha}{2}}I_{\left|\alpha\right|}\left(2 r_t M\sqrt{q_{l|k}^{\left(t\right)}q_{k|k}^{\left(t\right)}}\right) \right) \right).
\end{align} 
\hrulefill
\end{figure*}
where the right-hand side of \((a)\) in \eqref{eq:307} denotes the error rate assuming that the MVF always breaks ties in favor of an incorrect class. Moreover, \((b)\) is a result of Lemma~\ref{lemma:5}, and \((c)\) is a result of Lemma~\ref{lemma:4}. Please note that \(\mathbf{V}\) is Multinomial random vector and \(\textrm{Pr} \resizebox{.9 \hsize}{!}{\(\left({\bigcup_{t=1}^T}{\bigcup_{\underset{l\neq k}{l=1}}^K} \left\{\hat{V}_l^{(t)}-\hat{V}_k^{(t)} \geq E\left\{V_l^{(t)}-V_k^{(t)}\right\}+r_t M \delta_{l|k}\right\} \Big|X=x_k\right)\)}\) is a monotonic function of \(M\), hence it satisfies the conditions of Lemma \ref{lemma:4}. Next, \((d)\) is due to the independence of \(\hat{V}_l^{(t)}\)'s, \((d)\) is a result of Lemma \ref{lemma:5}, and finally, \((e)\) is derived by substituting \(\textrm{Pr}\left( \hat{V_l} - \hat{V_k} = \alpha | X=x_k\right)\) with the Skellam probability mass function \cite{Skellam}. 
\end{IEEEproof}

As we discussed before, a higher \(\delta_{l|k}\) is desirable as it shows a lower chance of outputting \(\hat{X}=x_l\) while \(X=x_k\). This is also evident from \eqref{eq:76} since a higher \(\delta_{l|k}\) increases the value of \(\alpha_0\) which in turn increases the lower limit of the inner summation. This in turn reduces the right-hand side of \eqref{eq:76} as the typical value of the inner summation is always positive. 

Next, we simplify the upper bound on the error rate of the MVF for i.non-i.d. voters in order to provide a better intuition about the behavior of \eqref{eq:76} versus the number of voters, \(M\), both for the asymptotic and the transient scenarios.

\subsection{Asymptotic Behaviour Of The Error Rate}
In the following, we first state a theorem and then exploit it to simplify the upper bound in~\eqref{eq:76} in the asymptotic scenario. 
\begin{theorem}\label{prop:10}
    \sloppy Assuming that \(\delta_{l_1,l_2|k}>0\) for given \(l_1, l_2 \text{~and~} k \in [K]\), we have
    \begin{equation}
        \lim_{M\to \infty} \textrm{Pr}\left(V_{l_1}<V_{l_2}|X=x_k\right)=0.
    \end{equation}
\end{theorem}
\begin{IEEEproof}
    Please refer to Appendix E.
\end{IEEEproof}

\begin{corollary}\label{cor:11}
    In the asymptotic scenario, where the number of voters goes to infinity, the error rate conditioned on \(X=x_k\) satisfies
    \begin{align}\label{eq:327}
    \lim_{M\to \infty} P_{e|x_k}&= \lim_{M\to \infty} \textrm{Pr}\left(\hat{X}=X \Big| X=x_k \right) \nonumber\\
                                &= \begin{cases}
                                    0  & \text{if~} \forall \text{~} l\neq k, \delta_{l|k}>0\\
                                    1  & \text{if~} \exists \text{~} l\neq k, \delta_{l|k}<0.\\
    \end{cases}
    \end{align}    
\end{corollary}
\begin{IEEEproof}
The proof is straightforward. Starting from the first case in \eqref{eq:327}, if \(\delta_{l|k}>0\), \(\forall \, l\neq k \in [K]\), then due to the Theorem \ref{prop:10}, we have 
\begin{equation}
    \lim_{M\to \infty} \textrm{Pr}\left(V_k<V_l|X=x_k\right)=0,\, \forall \, l\neq k \in [K].
\end{equation}
which implies that 
\begin{equation}
    \lim_{M\to \infty} P_{e|x_k} = 0. 
\end{equation}
On the contrary, if we have \(\delta_{l|k}<0\) for at least one \(l \neq k \in [K]\), then due to the Theorem \ref{prop:10}, we have
\begin{equation}
    \lim_{M\to \infty} \textrm{Pr}\left(V_l<V_k|X=x_k\right)=0,
\end{equation}
which implies that 
\begin{equation}
    \lim_{M\to \infty} P_{e|x_k} = 1. 
\end{equation}
\end{IEEEproof}

\begin{corollary}\label{cor:3}
    Let \(\mathcal{A}\) be the largest subset of \([K]\) such that \(\forall k \in \mathcal{A}\) we have \(\delta_{l|k}>0\), \(\forall l\in[K] \setminus k\). Then, the error rate of the MVF-based DFT converges towards \(1-\frac{|\mathcal{A}|}{K}\) as \(M\) grows.
\end{corollary}
\begin{IEEEproof}
    The proof is straightforward. From \eqref{eq:27}, we have
\begin{align}\label{eq:1111111}
    \lim_{M\to \infty}P_e &=\lim_{M\to \infty}\sum_{k=1}^K \textrm{Pr}\left(X=x_k\right) P_{e|x_k} \nonumber\\
                          &= \sum_{k\in\mathcal{A}}\frac{1}{K} \times 0+ \sum_{k\in [K]\backslash \mathcal{A}} \frac{1}{K}  \nonumber\\
                          &=1-\frac{|\mathcal{A}|}{K}.     
\end{align}
\end{IEEEproof}
Corollary \ref{cor:3} extends Corollary \ref{cor:2}, hence generalizing the Condorcet Jury Theorem even further. Due to the Corollary \ref{cor:3}, we need \(\delta_{l|k}>0\), \(\forall l, k \in [K]\), where \(l \neq k\), for the MVF to work reliably, i.e.,
\begin{align}\label{eq:501111}
    \lim_{M\to \infty}P_{e} = 0  \iff \forall l,k \in [K],\text{~where~} l\neq k, \text{~}\delta_{l|k}>0.
\end{align}

\subsection{Transient Behaviour Of The Error Rate}
Assuming \(\delta_{l|k}>0\), \(\forall l,k \in [K]\), where \(l \neq k\), we can derive an upper bound on the slope of decay of \(P_e\) versus \(M\) for i.non-i.d. voters, similar to what we had in Theorem \ref{theorem:11} for i.i.d. voters. We first define an auxiliary variable, \(\epsilon^*\), as
\begin{align}\label{eq:240}
    \epsilon^* \triangleq  \underset{k\neq l}{\underset{k,l,t}{\min}} \left(\frac{r_t \delta_{l|k}^2}{8\left(q_{l|k}^{(t)}+\frac{\delta_{l|k}}{6}\right)} \right).
\end{align}

\begin{theorem}\label{prop:3} 
If \(\delta_{l|k}>0\), \(\forall l,k \in [K]\), then we have
\begin{equation}\label{eq:800}
    P_e \leq 2KT \exp{\left(-M\epsilon^* \right)}.
\end{equation}
\end{theorem}
\begin{IEEEproof}
    Please refer to Appendix F.
\end{IEEEproof}
The upper bound derived in Theorem \ref{prop:3} is much looser than the upper bound obtained in Theorem \ref{theorem:4}, however, it is very insightful. The upper bound in \eqref{eq:800} shows that the error rate exponentially decays towards \(0\) as \(M\) grows since \(\epsilon^*\), defined in \eqref{eq:240}, increases linearly with \(M\). 
More specifically, using Theorem~\ref{prop:3} we can bound the slope of the decay of the error rate for the case where \(\delta_{l|k}>0\), \(\forall l,k \in [K]\), as
\begin{equation}\label{eq:801}
    \frac{\ln{P_e}}{M} \leq \frac{\ln{2KT}}{M}- \epsilon^* \to -\epsilon^*, \text{~as~} M \to \infty. 
\end{equation}
Therefore, if \(\delta_{l|k}>0\), \(\forall l,k \in [K]\), the error rate exponentially decays towards zero with a slope as most \(\frac{\ln{2KT}}{M}- \epsilon^*\),
as \(M \to \infty\), where for large values of \(M\) the upper bound on the slope is a constant number given by \(\epsilon^*=\underset{k\neq l}{\underset{k,l,t}{\min}} \left(\frac{r_t \delta_{l|k}^2}{8\left(q_{l|k}^{(t)}+\frac{\delta_{l|k}}{6}\right)} \right)\).

\section{The Majority Voting Function In Comparison With The Truth Discovery Algorithms}
The truth discovery algorithms are an important class of DFTs that combine the outputs of several weak classifiers/ voters without ever knowing the ground truth reliability of the classifiers/ voters. While the MVF assumes that all voters are equally reliable, truth discovery algorithms try to guess the reliability index of each voter from their past votes. 

In order to obtain a better understanding of the truth discovery algorithms, we will describe them briefly in the following. Let \(w_m^{(t)}\) denote the estimated reliability index of voter \(O_m\) at the \(t\)'th round of running the truth discovery algorithm.
The first time that a truth discovery algorithm is executed, when \(t=1\), the truth discovery algorithms often initialize the reliability index of all voters uniformly \cite{deducingtruth}, i.e., 
\begin{equation}\label{eq:1}
    w_m^{\left(1\right)}=c, \forall m \in [M].
\end{equation}
Once the first decision fusion task is completed, the reliability index of each voter is updated by minimizing a cost function. For instance, consider the cost function \(g\left(.,.\right)\) defined in \cite{deducingtruth}
\begin{equation}\label{eq:2}
    g\left(\mathbf{Y}^{(t)},\mathbf{w}^{(t)}\right)=\sum_{m=1}^{M} w_m^{(t)} d\left(Y_M^{(t)},\hat{X^{(t)}}\right),
\end{equation}
where \(\mathbf{w^{(t)}}=\left(w_1^{(t)},w_2^{(t)},\dots,w_M^{(t)}\right)\) is the reliability vector at the beginning of round \(t\), \(Y_m^{(t)}\) denotes the value reported by \(O_m\) at round \(t\), \(\mathbf{Y^{(t)}}=\left\{Y_1^{(t)}, Y_2^{(t)}, \dots, Y_M^{(t)}\right\}\), \(\hat{X}^{(t)}\) is the estimation of the decision fusion system about the unknown classification, and \(d\left(.,.\right)\) is a distance function defined by the system design.
The goal is to minimize the cost function by optimizing it with respect to \(\mathbf{w}^{(t)}\) and \(\hat{X}^{(t)}\) recursively. 

For a given \(k\in[K]\), we compare the reliability of the MVF with the truth discovery algorithms under two alternative assumptions: 

\subsubsection{If \(\forall l\neq k \in [K] \text{ we have }\delta_{l|k}>0\)}
We showed in Section III that if \(\delta_{l|k}>0\), \(\forall l\neq k \in [K]\), as we increase the number of independent voters, the MVF can obtain an arbitrarily small error rate in detecting \(X\) when \(X=x_k\). In this scenario, in the best case situation, the truth discovery algorithms may perform even better than the MVF as their performance continues to improve not only by increasing the number of voters but also with time. This is because when \(\delta_{l|k}>0\), \(\forall l\neq k \in [K]\), there is a high probability that the final output of the decision unit, \(\hat{X}\), at time \(t\), is correct which in turn results in increasing the accuracy of estimation of the reliability index of the voters. Hence, over time the more reliable voters will attract a higher reliability index and help in increasing the accuracy of the decision fusion system. However, this is the best-case situation. There are also circumstances where implementing truth discovery algorithms may result in highly erroneous outcomes even if \(\delta_{l|k}>0\), \(\forall l\neq k \in [K]\). This can occur because there is always a non-zero probability that during the first few rounds of the operation of the decision fusion system, the truth discovery algorithm outputs incorrect values, e.g., due to the inherent noise of the data sources and the processing algorithms that the voters use to prepare their votes. In spite of the lack of ground truth in the system, the truth discovery algorithms often interpret the final output of the system as the ground truth. As a result, the reliability index of the erroneous voters that have voted for the classes that are deemed as correct classes will be increased. Consequently, the error rate of the system increases and may lead to the domination of unreliable voters in the system. 

\subsubsection {If \(\exists l\neq k \in [K] \text{ such that }\delta_{l|k}<0\)}
Recalling our results in Section III, if \(\exists l\neq k \in [K] \text{ such that }\delta_{l|k}<0\), then the MVF will have a high error rate in detecting \(X\) when \(X=x_k\). In fact, we showed that in this scenario the error rate increases exponentially as we increase the number of independent voters. In this scenario, the performance of the truth discovery algorithms cannot be better than the MVF. From \eqref{eq:1}, we observe that at time \(t=1\), the output of a typical truth discovery-based DFT is equal to the output of the MVF. If there is at least one \(l\in[K]\) such that \(\delta_{l|k}<0\), whether we employ the  MVF or the truth discovery algorithms, the output of the data fusion system is erroneous with a high probability\footnote{Some truth discovery implementation may use other statistics like median or just randomly pick a value as the output. However, in the absence of historical data, none of these alternative methods is more reliable than the MVF in obtaining the first output of the system}. 
However, in the truth discovery algorithms, those voters that have mistakenly reported the same value as the majority of the voters reported in the first round will receive a higher reliability index at the beginning of the second round. Therefore, those unreliable voters will have a higher influence on the data fusion system in the second round. 

In other words, the idea of guessing the reliability of the voters from their past votes without having access to the past ground truths, and then, combining those past votes to generate reliability indexes, creates a positive feedback loop in the system. When compared to the MVF, the positive feedback loop of the truth discovery algorithms in cases when \(\delta_{l|k}>0\), \(\forall l\neq k \in [K]\), amplifies the accuracy, in the best-case scenario. However, the same positive feedback loop of the truth discovery algorithms in cases when \(\delta_{l|k}<0\), for at least one \(l\neq k \in [K]\), amplifies the error rate. In other words, in terms of their error rate, the truth discovery algorithms, in the best case, are simply a MVF with an amplification factor. 

Furthermore, the MVF is faster and simpler than the truth discovery algorithms, and since it does not require past votes, it is practical for a wider range of applications.
For instance, truth discovery algorithms are not applicable if a decision fusion system requires the voters to join the network anonymously in order to preserve their privacy and neutralize bribing attacks \cite{PrivacyDataFeed}. Also, in some applications, the voters may rarely participate in a task (e.g., when reporting an incident in a system) or they may prefer to remain anonymous for security and privacy reasons (e.g., when reporting corruption in an organization), and as a result, zero or little information about their previously reported values is known to the decision unit.

Hence, employing the MVF as a DFT could be a better choice especially when a large number of independent voters are available in the system.

\section{Numerical Results}
We utilize the Human Activity Recognition Trondheim (HARTH) data set \cite{HARTHdataSet} for our numerical analysis of the error rate of the MVF. To this end, we first train $1000$ independent k-nearest neighbors (k-NN) classifiers \cite{kNN} using this data set and then employ the MVF to combine the outputs of a randomly selected subset of them. We use the error rate of the output of the MVF to evaluate the accuracy of our upper bound derived in Theorem~\ref{theorem:1} and Theorem~\ref{theorem:4}. It is essential to emphasize that evaluating the performance of the base classifiers is not within the scope of our current focus. 

The HARTH data set comprises over $6$ million time series sample readings from two three-axial accelerometers attached to the thigh and lower back of $22$ participants during their regular working hours. The samples are annotated with $12$ different labels to distinguish activities, such as stairs (ascending), stairs (descending), shuffling (standing with leg movement), cycling (standing), cycling (sitting), transport (sitting), and transport (standing), using video recordings from a chest-mounted camera \cite{HARTHdataSet}.

The preprocessing of the data set involved removing samples associated with two classes, cycling (standing) and cycling (sitting), due to low sample frequency. We also balanced the number of samples across the remaining $10$ labels by randomly removing samples from classes with a high number of samples. Each original sample with $6$ attributes (obtained from accelerometer readings) is supplemented with an additional $6$ attributes representing the differences between the readings of each sample with the sample taken exactly $5$ time slots later, capturing accelerometer changes due to movements. We further remove some samples with low annotation accuracy, such as those at the edge of activity transitions. Finally, we randomly select $80\%$ of the samples for training our classifiers, with the remaining serving as the test data set.

In the following, we present our numerical analysis in three subsections. The first two subsections focus on numerical results using the HARTH data set, simulating i.i.d. and i.non-i.d. voters, respectively. In the third subsection, we provide numerical results based on a generated data set, aiming to explore the performance of our upper bound for i.i.d. voters across various scenarios in greater detail.

\subsection{Numerical Analysis Assuming I.I.D. Voters}
We create $1000$ 3-NN classifiers, each trained on separate subsets of randomly selected data points to guarantee the independence of classifiers. To ensure the identical distribution of classifiers' outputs, each classifier is trained with an equal number of data points from every class. Test results demonstrate a negligible variance among probability transition matrices of different classifiers, thereby supporting the assumption of i.i.d. outputs. To simulate the error rate of the MVF for i.i.d. voters, we randomly select $M$ classifiers to classify the data points in the test data set and then compare the output of MVF with the true label of the data point.

To calculate the upper bound, we use the probability transition matrix of the worst-performing classifier, given by
\begin{equation}\label{eq:924}
    \mathbf{P}=\resizebox{.93 \hsize}{!}{$
\begin{bmatrix}
0.33 & 0.03 & 0.06 & 0.30 & 0.18 & 0.01 & 0.00 & 0.00 & 0.04 & 0.06 \\
0.04 & 0.87 & 0.00 & 0.03 & 0.03 & 0.00 & 0.00 & 0.00 & 0.00 & 0.02 \\
0.07 & 0.00 & 0.56 & 0.18 & 0.01 & 0.16 & 0.00 & 0.00 & 0.01 & 0.00 \\
0.07 & 0.00 & 0.10 & 0.67 & 0.04 & 0.03 & 0.00 & 0.00 & 0.05 & 0.04 \\
0.16 & 0.03 & 0.07 & 0.24 & 0.36 & 0.01 & 0.00 & 0.00 & 0.08 & 0.05 \\
0.02 & 0.00 & 0.11 & 0.03 & 0.02 & 0.82 & 0.00 & 0.00 & 0.00 & 0.00 \\
0.00 & 0.00 & 0.00 & 0.00 & 0.00 & 0.00 & 0.99 & 0.01 & 0.00 & 0.00 \\
0.00 & 0.00 & 0.00 & 0.00 & 0.00 & 0.00 & 0.06 & 0.94 & 0.00 & 0.00 \\
0.03 & 0.00 & 0.01 & 0.05 & 0.03 & 0.00 & 0.00 & 0.00 & 0.81 & 0.05 \\
0.07 & 0.00 & 0.04 & 0.11 & 0.02 & 0.01 & 0.00 & 0.00 & 0.13 & 0.60 \\
\end{bmatrix}$
}
\end{equation}

\begin{figure} 
\centering
\includegraphics[width=.55\textwidth]{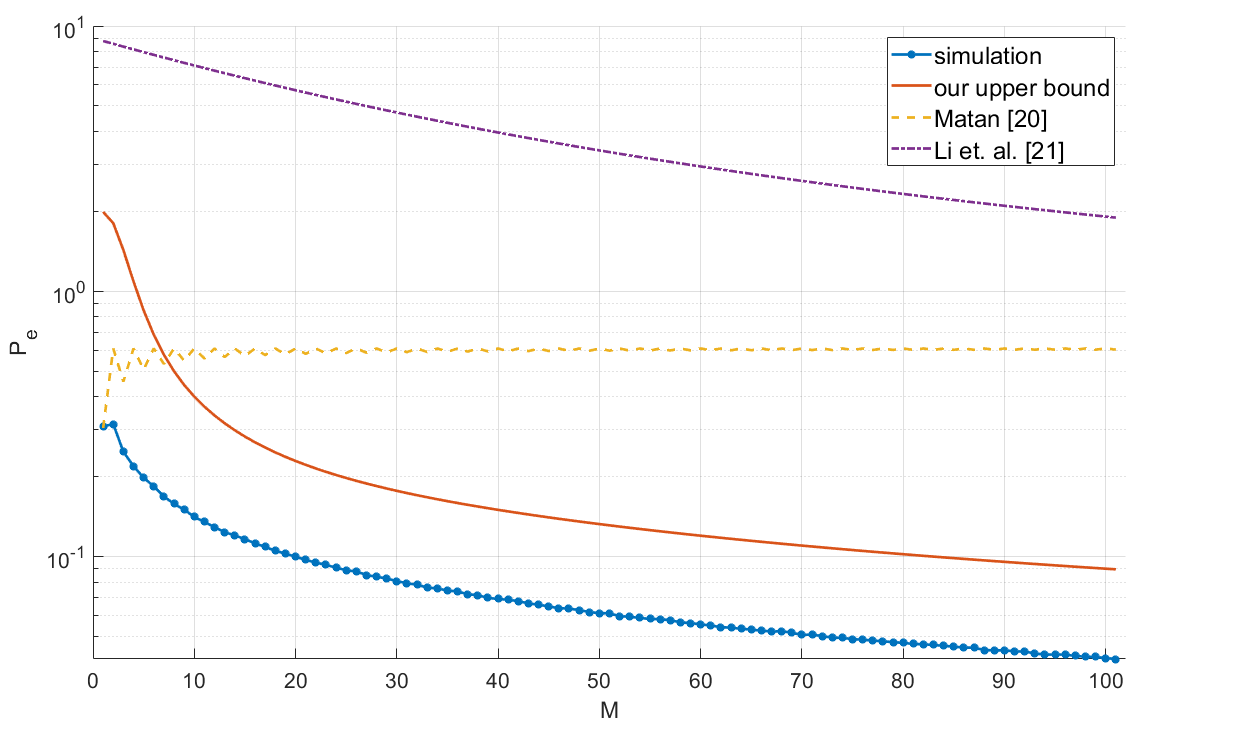}
\caption{Our theoretical upper bound on the error rate of the majority voting function is plotted and compared with simulation results and two other upper bounds derived in \cite{Matan96onvoting} and \cite{li2014error}.
}\label{fig:UpperboundIID}
\end{figure}

We expect the error rate to decay towards zero as $M$ increases since $p_{k|k}>p_{l|k}$, $\forall l\neq k$ in \eqref{eq:924}. Fig.~\ref{fig:UpperboundIID} illustrates the error rate of the MVF as a function of the number of independent classifiers voting their opinion. This figure compares the simulation results with the upper bound on the error rate of the MVF, presented in Theorem \ref{theorem:1}, along with the upper bounds derived in \cite{li2014error} and \cite{Matan96onvoting}.

As shown in Fig.~\ref{fig:UpperboundIID}, our upper bound in Theorem \ref{theorem:1} consistently remains at a close distance from the true error rates obtained through simulations\footnote{For a small number of voters, our upper bound may exceed one. Note that we can always use one as a trivial upper bound for any probability value. However, when $M$ increases, our upper bound decays exponentially towards zero.}. In contrast, the upper bound established in \cite{li2014error} maintains a significantly larger gap and diverges from the actual error rate as the number of voters, \(M\), increases. Moreover, the upper bound derived in \cite{Matan96onvoting} does not exhibit a decrease as the number of independent voters grows.

\subsection{Numerical Analysis Assuming I.Non-I.D. Voters}
Following a similar approach as employed for simulating i.i.d. voters, we employ the training data set to create 1000 independent classifiers. However, in order to generate non-identically distributed classifiers, we adopt distinct classification algorithms and sensor data, to create four classifier groups as detailed below:
\begin{itemize}
    \item $\mathcal{G}_1$: 250 classifiers use the 2-NN classification algorithm and are trained on attributes corresponding to the sensors attached to the thigh of the participants.
    \item $\mathcal{G}_2$: 250 classifiers use the 3-NN classification algorithm and are trained on attributes corresponding to the sensors attached to the thigh of the participants.
    \item $\mathcal{G}_3$: 250 classifiers use the 2-NN classification algorithm and are trained on attributes corresponding to the sensors attached to the lower back of the participants.
    \item $\mathcal{G}_4$: 250 classifiers use the 3-NN classification algorithm and are trained on attributes corresponding to the sensors attached to the lower back of the participants.
\end{itemize}
To simulate the error rate of the MVF, we select $M$ classifiers at random where it is equally likely for each classifier to be selected from $\mathcal{G}_1$, $\mathcal{G}_2$, $\mathcal{G}_3$, or $\mathcal{G}_4$. Next, we evaluate the classifiers using the test data set and aggregate their outputs using the MVF, enabling a comparison between the resultant outcomes and the actual labels of the data points.

\begin{figure} 
\centering
\includegraphics[width=.55\textwidth]{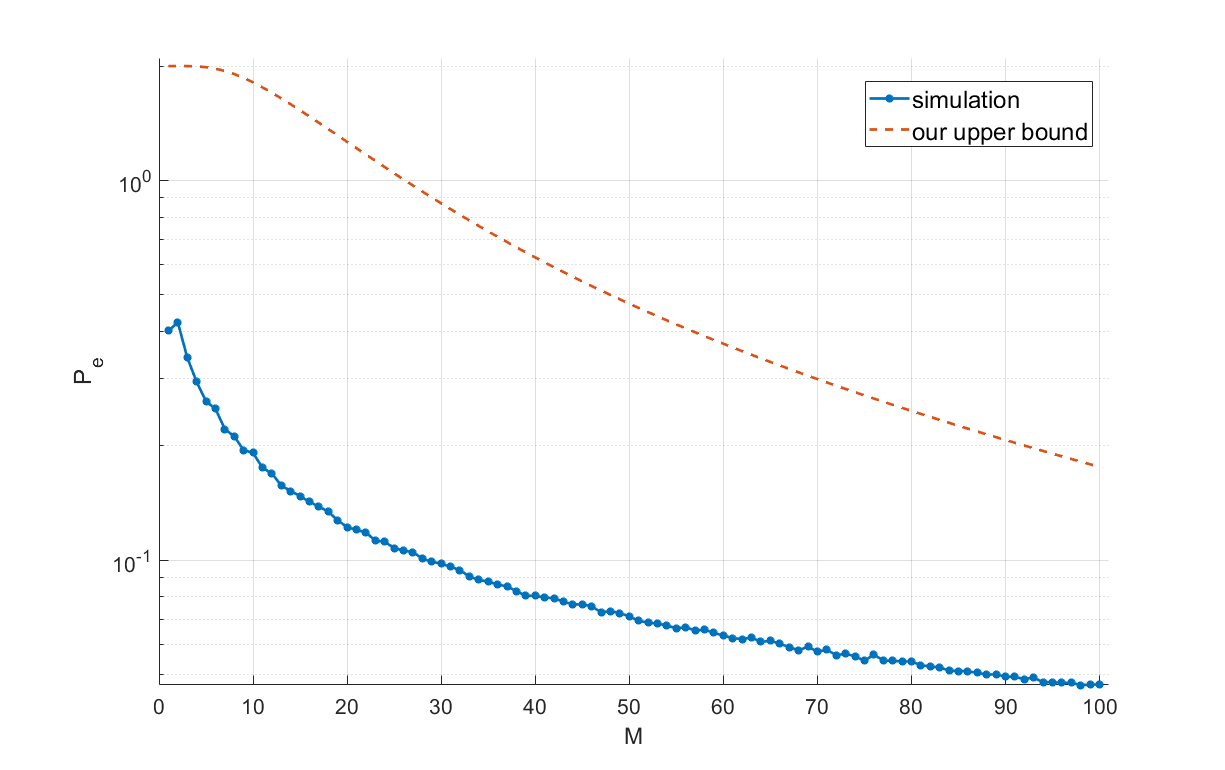}
\caption {
Our upper bound on the error rate of the MVF for i.non-i.d. voters is illustrated and compared to the actual error rate of $M$ independent voters each drawn at random from one of the four groups of classifiers with probability transition matrices in \eqref{eq:925}.}
\label{fig:UpperboundINonID}
\end{figure}

To test the accuracy of our upper bound, we first estimate  the probability transition matrices corresponding to each group of classifiers using the test data set as 
\begin{align}\label{eq:925}
    \mathbf{P}^{(1)}&=\resizebox{.868 \hsize}{!}{$\begin{bmatrix}
0.44 & 0.04 & 0.04 & 0.16 & 0.23 & 0.01 & 0.00 & 0.00 & 0.03 & 0.05 \\
0.05 & 0.80 & 0.00 & 0.02 & 0.09 & 0.00 & 0.00 & 0.00 & 0.00 & 0.04 \\
0.10 & 0.00 & 0.41 & 0.07 & 0.04 & 0.22 & 0.10 & 0.00 & 0.04 & 0.01 \\
0.15 & 0.01 & 0.07 & 0.41 & 0.13 & 0.03 & 0.02 & 0.00 & 0.15 & 0.05 \\
0.22 & 0.06 & 0.06 & 0.13 & 0.46 & 0.01 & 0.00 & 0.00 & 0.03 & 0.03 \\
0.02 & 0.00 & 0.15 & 0.01 & 0.00 & 0.48 & 0.31 & 0.00 & 0.01 & 0.01 \\
0.01 & 0.00 & 0.08 & 0.01 & 0.00 & 0.27 & 0.59 & 0.00 & 0.02 & 0.01 \\
0.00 & 0.00 & 0.00 & 0.00 & 0.00 & 0.00 & 0.00 & 0.99 & 0.00 & 0.00 \\
0.05 & 0.00 & 0.06 & 0.16 & 0.04 & 0.03 & 0.03 & 0.00 & 0.50 & 0.13 \\
0.06 & 0.02 & 0.01 & 0.06 & 0.05 & 0.00 & 0.01 & 0.00 & 0.11 & 0.67 \\
\end{bmatrix}
$} \nonumber\\ 
\mathbf{P}^{(2)}&=\resizebox{.855 \hsize}{!}{$\begin{bmatrix}
0.24 & 0.08 & 0.08 & 0.15 & 0.19 & 0.01 & 0.00 & 0.00 & 0.10 & 0.15 \\
0.10 & 0.61 & 0.01 & 0.06 & 0.10 & 0.00 & 0.00 & 0.00 & 0.05 & 0.06 \\
0.14 & 0.00 & 0.38 & 0.15 & 0.08 & 0.17 & 0.00 & 0.00 & 0.03 & 0.06 \\
0.13 & 0.01 & 0.15 & 0.30 & 0.12 & 0.05 & 0.00 & 0.00 & 0.07 & 0.17 \\
0.17 & 0.06 & 0.08 & 0.14 & 0.26 & 0.01 & 0.00 & 0.00 & 0.13 & 0.15 \\
0.02 & 0.00 & 0.19 & 0.02 & 0.01 & 0.75 & 0.00 & 0.00 & 0.01 & 0.01 \\
0.00 & 0.00 & 0.00 & 0.00 & 0.00 & 0.01 & 0.85 & 0.14 & 0.00 & 0.00 \\
0.00 & 0.00 & 0.00 & 0.00 & 0.00 & 0.00 & 0.14 & 0.86 & 0.00 & 0.00 \\
0.07 & 0.02 & 0.04 & 0.08 & 0.10 & 0.01 & 0.01 & 0.00 & 0.51 & 0.15 \\
0.13 & 0.04 & 0.06 & 0.17 & 0.15 & 0.01 & 0.00 & 0.00 & 0.12 & 0.32 \\
\end{bmatrix}$}\nonumber\\
\mathbf{P}^{(3)}&=\resizebox{.855 \hsize}{!}{$\begin{bmatrix}
0.46 & 0.04 & 0.03 & 0.15 & 0.23 & 0.01 & 0.00 & 0.00 & 0.03 & 0.06 \\
0.04 & 0.81 & 0.00 & 0.01 & 0.08 & 0.00 & 0.00 & 0.01 & 0.00 & 0.04 \\
0.08 & 0.00 & 0.44 & 0.06 & 0.04 & 0.24 & 0.10 & 0.00 & 0.04 & 0.01 \\
0.14 & 0.01 & 0.07 & 0.42 & 0.13 & 0.02 & 0.02 & 0.00 & 0.16 & 0.04 \\
0.21 & 0.05 & 0.05 & 0.12 & 0.49 & 0.01 & 0.00 & 0.00 & 0.03 & 0.03 \\
0.01 & 0.00 & 0.13 & 0.01 & 0.00 & 0.51 & 0.31 & 0.00 & 0.01 & 0.00 \\
0.01 & 0.00 & 0.08 & 0.01 & 0.00 & 0.27 & 0.59 & 0.00 & 0.02 & 0.01 \\
0.00 & 0.00 & 0.00 & 0.00 & 0.00 & 0.00 & 0.00 & 0.99 & 0.00 & 0.00 \\
0.04 & 0.00 & 0.06 & 0.14 & 0.03 & 0.03 & 0.04 & 0.00 & 0.56 & 0.11 \\
0.06 & 0.02 & 0.01 & 0.06 & 0.05 & 0.00 & 0.01 & 0.00 & 0.12 & 0.68 \\
\end{bmatrix}$} \nonumber\\
\mathbf{P}^{(4)}&=\resizebox{.855 \hsize}{!}{$
\begin{bmatrix}
0.24 & 0.07 & 0.08 & 0.14 & 0.19 & 0.01 & 0.00 & 0.00 & 0.10 & 0.16 \\
0.09 & 0.64 & 0.01 & 0.04 & 0.10 & 0.00 & 0.00 & 0.00 & 0.05 & 0.06 \\
0.12 & 0.00 & 0.41 & 0.13 & 0.07 & 0.17 & 0.00 & 0.00 & 0.03 & 0.06 \\
0.12 & 0.01 & 0.15 & 0.30 & 0.11 & 0.06 & 0.00 & 0.00 & 0.06 & 0.17 \\
0.16 & 0.06 & 0.07 & 0.14 & 0.26 & 0.01 & 0.00 & 0.00 & 0.13 & 0.16 \\
0.01 & 0.00 & 0.14 & 0.01 & 0.00 & 0.82 & 0.00 & 0.00 & 0.00 & 0.01 \\
0.00 & 0.00 & 0.00 & 0.00 & 0.00 & 0.00 & 0.88 & 0.12 & 0.00 & 0.00 \\
0.00 & 0.00 & 0.00 & 0.00 & 0.00 & 0.00 & 0.19 & 0.81 & 0.00 & 0.00 \\
0.06 & 0.02 & 0.04 & 0.08 & 0.10 & 0.01 & 0.01 & 0.01 & 0.53 & 0.15 \\
0.12 & 0.04 & 0.06 & 0.17 & 0.14 & 0.01 & 0.00 & 0.00 & 0.12 & 0.35 \\
\end{bmatrix}$}, 
\end{align}
where $\mathbf{P}^{(1)}, \mathbf{P}^{(2)}, \mathbf{P}^{(3)}$, and $\mathbf{P}^{(4)}$ are the probability transition matrices corresponding to $\mathcal{G}_1$, $\mathcal{G}_2$, $\mathcal{G}_3$, and $\mathcal{G}_4$, respectively. Subsequently, we employ Theorem \ref{theorem:4} to find the upper bound of the MVF aggregating $M$ i.non-i.d. voters drawn uniformly at random\footnote{While we have chosen to use a uniform distribution for this simulation, it's important to stress that this choice is made without loss of generality; our system model in Section IV encompasses scenarios where voter proportions in each category need not be equal.} from $\mathcal{G}_1$, $\mathcal{G}_2$, $\mathcal{G}_3$, or $\mathcal{G}_4$. The numerical results are presented in Fig.~\ref{fig:UpperboundINonID}.

Using \eqref{eq:301}, we realize that $\delta_{l|k}>0$ for all $l,k \in [10]$, with its minimum value being $0.135$ for $k=1$ and $l=5$. Therefore, due to Corollary \ref{cor:3}, we expect the error rate to go towards zero as $M$ increases which is in agreement with the simulation results in Fig. \ref{fig:UpperboundINonID}. Additionally, we calculate $\epsilon^*$ according to \eqref{eq:240} and employ \eqref{eq:801} to obtain $-5.62 \times 10^{-4}$ as an upper bound on the slope of the decay of error rate as $M$ grows to infinity. While this is a weak upper bound, especially when compared with our upper bound on the error rate itself, it is useful for demonstrating that the error rate is decaying exponentially with $M$.

\subsection{Extended Numerical Analysis with Generated Data for I.I.D. Voters}
In this subsection, we further explore the performance of the MVF assuming i.i.d. voters.
We simulate a \(10\)-class classification problem, i.e., \(|\mathcal{X}|=K=10\) where the probability of misclassification is uniform over all input and output classes in \(\mathcal{X}\). Let \(\textrm{Pr}\left( Y_m=x_k | X= x_k \right)= \gamma+\frac{1-\gamma}{K}\) and \(\textrm{Pr}\left( Y_m=x_l | X= x_k \right)= \frac{1-\gamma}{K}\). Hence, the probability transition matrix for the voters has the following format
\begin{align}\label{eq:923}
    \textrm{p}=\begin{bmatrix} 
                    \left(\gamma+\frac{1-\gamma}{K}\right) & \frac{1-\gamma}{K} &\dots &\frac{1-\gamma}{K} \\
                    \frac{1-\gamma}{K} &\left(\gamma+\frac{1-\gamma}{K}\right) &\dots &\frac{1-\gamma}{K} \\
                     \vdots &\vdots  &\ddots    &\vdots \\
                    \frac{1-\gamma}{K} &\frac{1-\gamma}{K}   &\dots   &\left(\gamma+\frac{1-\gamma}{K}\right)
                \end{bmatrix}, 
\end{align}
where \(\gamma \in [0,1]\) is the parameter we use to model the average level of accuracy of the voters. This is a standard way for modeling voters in a decision fusion system that is commonly known as the homogeneous Dawid-Skene model \cite{li2014error}. 
Using this model, we can study the MVF's error rate versus the accuracy of individual voters.
For instance, if \(\gamma=0\), we have the special case when we have zero accuracy and the expected number of votes submitted for different classes is uniformly distributed regardless of \(X\), while for \(\gamma=1\), we have full accuracy since all of the voters vote for the correct class. 

\begin{figure} 
\centering
\includegraphics[width=.4345\textwidth]{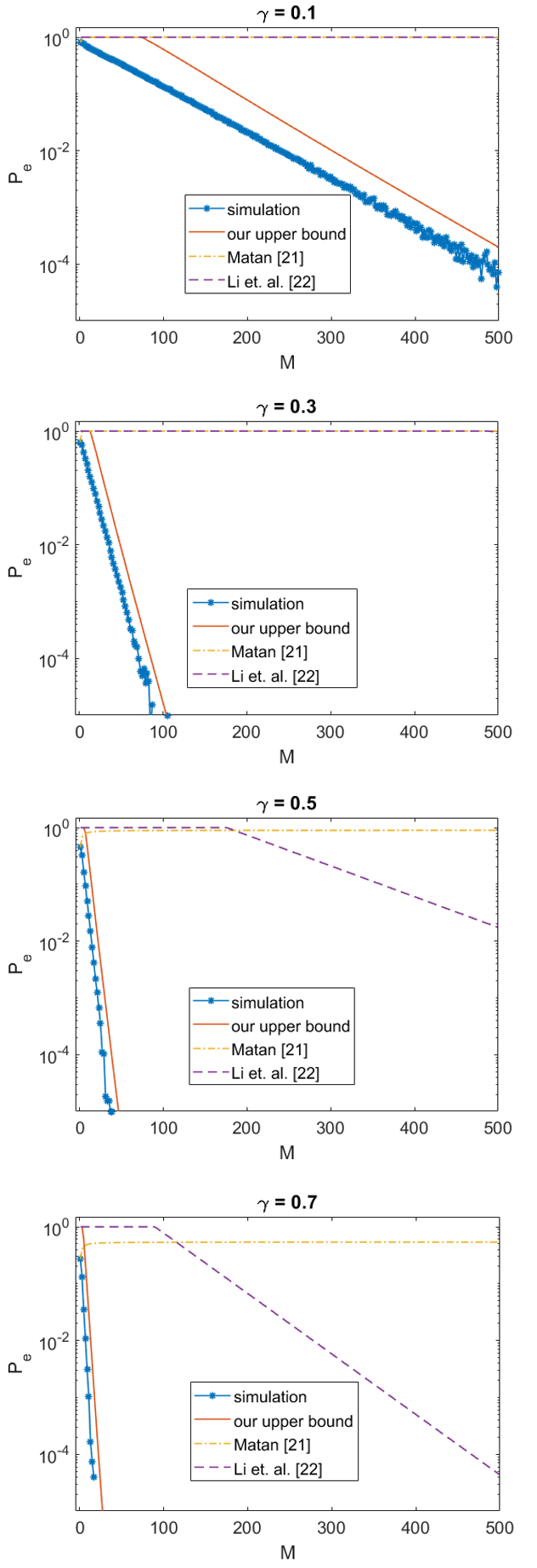}
\caption{Our theoretical upper bound on the error rate of the majority voting function is plotted for four different values of \(\gamma\) and compared with simulations results and two other upper bounds derived in \cite{li2014error} and \cite{Matan96onvoting}.
}\label{fig:UpperAndLowerBound}
\end{figure}

The performance of our upper bound on the error rate given in Theorem \ref{theorem:1}, is illustrated in Fig.~\ref{fig:UpperAndLowerBound} and is compared to the true error rate obtained by simulations. As we can see in Fig.~\ref{fig:UpperAndLowerBound}, the error rate of the MVF decreases exponentially as the number of voters increases. 
Assuming that the individual voters have an accuracy parameter \(\gamma=0.3\), a single voter makes a classification error with a probability as high as \(0.63\). However, our simulation shows that around \(30\) independent voters together with a MVF, can achieve an error rate less than \(10^{-2}\).  
This number reduces to as low as \(15\) voters if the accuracy parameter is as high as \(\gamma=0.5\).

In Fig.~\ref{fig:UpperAndLowerBound}, the upper bound we obtained in Theorem \ref{theorem:1} is compared to two other upper bounds previously derived in \cite{li2014error, Matan96onvoting} as well. 
The simulation results show that our upper bound keeps a constant distance from the true error rate obtained by simulations, while the upper bound obtained in \cite{li2014error} diverges from the true error rate as the number of voters, \(M\), increases and the upper bound derived in \cite{Matan96onvoting} does not decrease as the number of independent voters increases.

\begin{figure} 
\centering
\includegraphics[width=.53\textwidth]{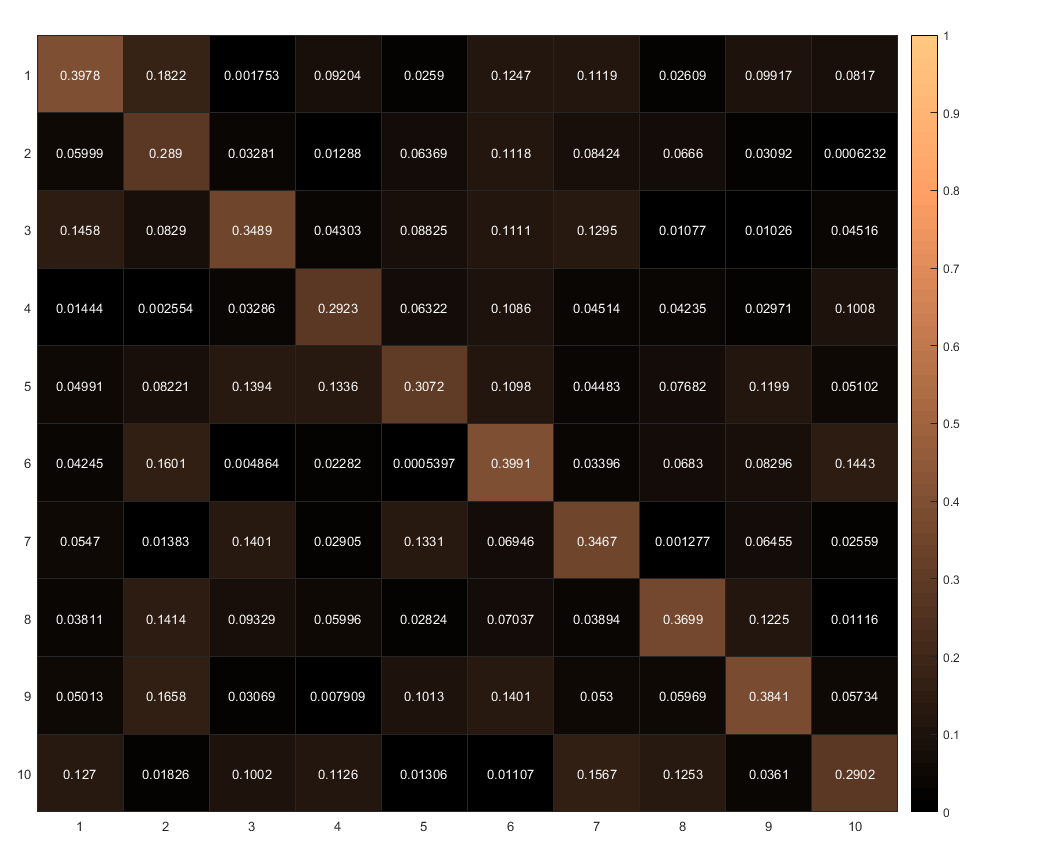}
\caption{The probability transition matrix of \(M\) i.i.d. voters is illustrated. This probability transition matrix is generated randomly and satisfies the right-hand side of \eqref{eq:501}.}\label{fig:heatmap}
\end{figure}

Next, we explore the performance of the upper bound we derived on the slope of decay of the error rate for the MVF. To this end, we first generate a random probability transition matrix that satisfies the right-hand-side of \eqref{eq:501} to model the probability transition matrices of \(M\) i.i.d. voters (Fig. \ref{fig:heatmap}).

\begin{figure} 
\centering
\includegraphics[width=.5\textwidth]{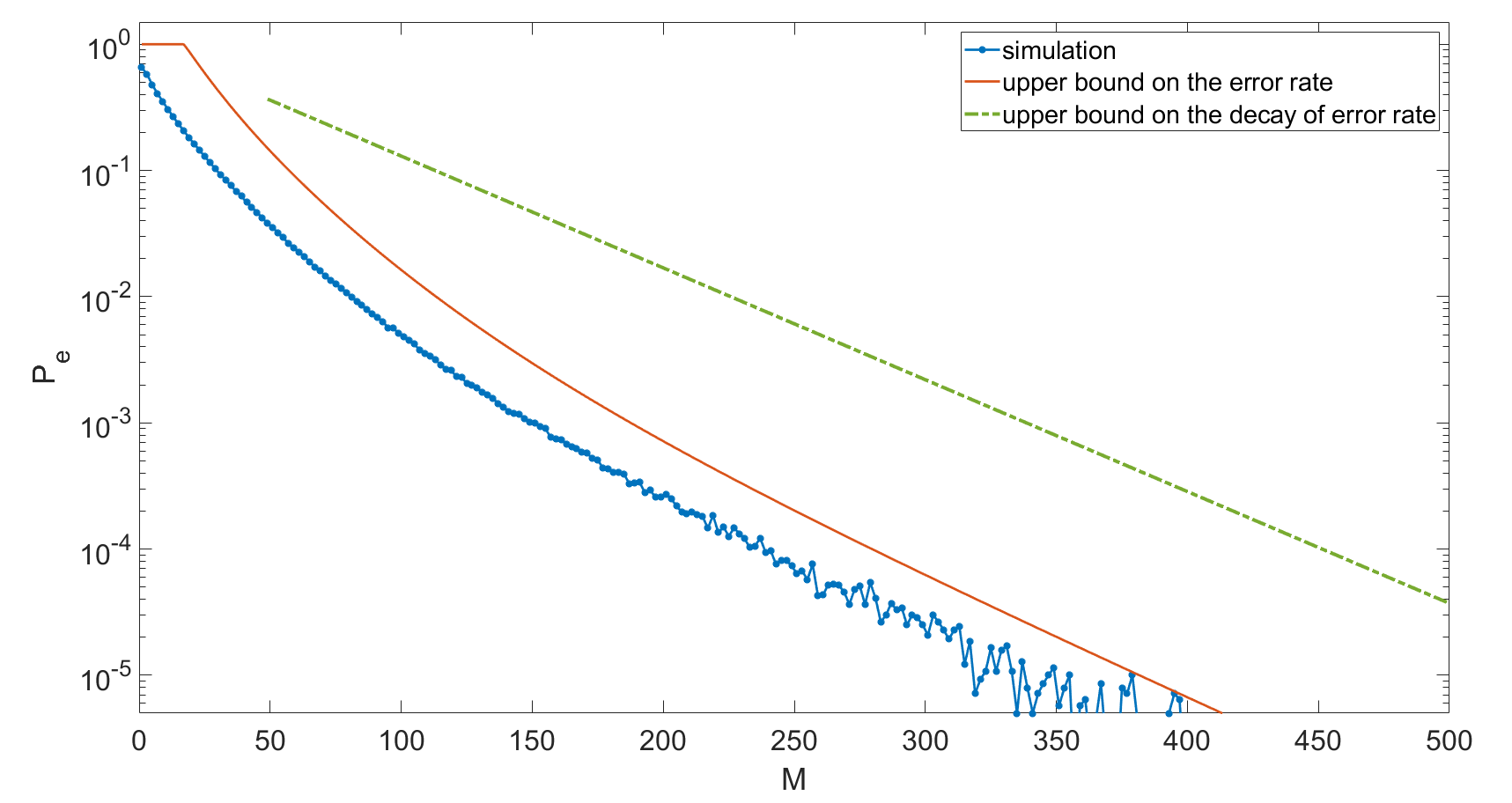}
\caption{The simulation result, the upper bound, and the upper bound on the slope of decay of the error rate for the MVF are illustrated.
}\label{fig:decay}
\end{figure}

Applying \eqref{eq:901} to the probability transition matrix in Fig. \ref{fig:heatmap}, we have
\begin{align*}
    \lim_{M\to \infty}\frac{\ln{P_e}}{M} \leq -\left(\sqrt{p_{k^*|k^*}}-\sqrt{p_{l^*|k^*}} \right)^2 +   \lim_{M\to \infty} \epsilon = -0.0204, 
\end{align*}
where $\left(k^*, l^*\right)= \left(10,7 \right)$. Hence, we expect the error rate to exponentially decay toward \(0\) as \(M\) grows, where the slope of the decay is at least \(-0.0204\). Fig. \ref{fig:decay} illustrates our upper bound of the error rate as well as our upper bound on the slope of decay of the error rate. 

\section{Conclusion}
We explored the accuracy of the majority voting function for the general multi-class classification problem. Leveraging techniques from the balls into bins problem, we established upper bounds for the error rates of the majority voting function, considering both i.i.d. and i.non-i.d. voters. Furthermore, we scrutinized the behavior of these bounds in both asymptotic and transient scenarios, obtaining the sufficient and necessary conditions for the error rate of the majority voting function to decay exponentially as the number of voters increases.
Building on these findings, we conducted an insightful analysis of truth discovery algorithms. We demonstrated that, in optimal scenarios, truth discovery algorithms essentially function as amplifications of the majority voting function, achieving low error rates only when majority voting does and vice versa. Conversely, in worst-case situations, truth discovery algorithms may exhibit higher error rates compared to the majority voting function.
Finally, we presented numerical results, demonstrating that our upper bounds closely align with the actual error rate of the majority voting function, as determined through numerical simulations. These insights collectively enhance our comprehension of the performance and accuracy of majority voting and truth discovery algorithms in multi-class classification settings, providing valuable guidance for estimating their accuracy in real-world applications.

\bibliographystyle{IEEEtran}
\bibliography{ref}

\appendices

\section{Proof Of Lemma \ref{lemma:4}}
In order to prove Lemma \ref{lemma:4}, we use a bound proposed in Theorem 5.10 in \cite{mitzenmacher}. This bound is provided below as a lemma for completeness. 
\begin{lemma} \label{lemma:3}
Let \(\mathbf{V}=\left(V_1,V_2,\dots,V_K\right)\) be a random vector with the following Multinomial distribution 
\begin{equation}\label{eq:13}
    \textrm{Pr}\left(\mathbf{V}=\mathbf{v}|X=x_k\right)= \frac{M!}{v_1!v_2!\dots v_K!}\times p_{1|k}^{v_1} \times p_{2|k}^{v_2} \times \dots \times p_{K|k}^{v_K},
\end{equation}
such that \(\sum_{k=1}^K v_k =M\) and 
\begin{equation}\label{eq:1011}
    p_{1|k}=p_{2|k}=\dots=p_{K|k}=\frac{1}{K},\, \forall k \in [K].
\end{equation}
Moreover, let \(\mathbf{\hat{V}}=\left(\hat{V_1},\hat{V_2},\dots,\hat{V_K}\right)\) be a vector of independent and identically distributed Poisson RVs with parameter \(\lambda=\frac{M}{K}\).
Finally, let \(\mathcal{E}\left(\mathbf{V}\right)\) be an event imposed on the Multinomial random vector \(\mathbf{V}\) whose probability is either monotonically increasing or monotonically decreasing in the number of voters, \(M\), and let \(\mathcal{E}\left(\hat{\mathbf{V}}\right)\) denote the same event applied on the random Poisson vector \(\hat{\mathbf{V}}\). Then, we have 
\begin{equation}
    \textrm{Pr}\left(\mathcal{E}\left(\mathbf{V}\right) \right) \leq 2 \textrm{Pr}\left(\mathcal{E}\left(\hat{\mathbf{V}}\right) \right).
\end{equation}
\end{lemma}

\begin{IEEEproof} The proof\footnote{The original Theorem has been stated for the balls and bins problem. However, we presented Lemma \ref{lemma:3} for voters and classes in a classification problem in order to keep our terminology consistent.} is provided in \cite{mitzenmacher}.
\end{IEEEproof}

In Lemma \ref{lemma:3}, it is assumed that \eqref{eq:1011} holds. In order to prove Lemma \ref{lemma:4}, we need to expand Lemma \ref{lemma:3} to the more general distribution of \(p_{l|k}\)'s where the conditional probabilities are not necessarily as in \eqref{eq:1011}, i.e., uniformly distributed. To this end, we first define the set \(\mathcal{Z}=\left\{z_1,z_2,\dots,z_R\right\}\) where \(R\to \infty\). Next, for each \(x_k\) we define the many-to-one mapping, \(f_k:\mathcal{Z}\to \mathcal{X}\), as
\begin{equation}
    f_k\left(z_r\right)= \begin{cases}
                        x_1 & r\leq R p_{1|k}, \\
                        x_l & R\sum_{i=1}^{l-1} p_{i|k} <r \leq R\sum_{i=1}^{l}p_{i|k}.
                        \end{cases}
\end{equation}
In words, using \(f_k\), we group the first \(R p_{1|k}\) elements in \(\mathcal{Z}\) and map it to class \(x_1\). Similarly, we group the next \(R p_{2|k}\) elements in \(\mathcal{Z}\) and map it to class \(x_2\), and so on until the last \(R p_{K|k}\) elements in \(\mathcal{Z}\) that are mapped to the last class in \(\mathcal{X}\), i.e., \(x_K\).

Given that \(X=x_k\), let \(g_k(.)\) be the random oracle that reads any submitted vote, \(Y_m\), and maps it to an element in its pre-image, \(f_k^{-1}\left(Y_m\right)\), uniformly at random. Moreover, let \(\mathbf{W}=\left(W_1, W_2, \dots, W_R \right)\), where \(W_r\) denotes the number of votes mapped to \(z_r\). Then, we have 
\begin{equation}\label{eq:1015}
    \textrm{Pr}\left(\mathbf{W}=\mathbf{w}|X=x_k\right)= \frac{M!}{w_1!w_2!\dots w_R!} \prod_{r=1}^R \left(\frac{1}{R}\right)^{w_r} ,   
\end{equation}
where \(\sum_{r=1}^R w_r= M\). This is similar to what we had in Lemma \ref{lemma:3} and is a result of the fact that the probability of any randomly selected vote gets mapped to \(z_r\) is equal for all \(r\in [R]\), i.e., 
\begin{equation}
    \textrm{Pr}\left(g_k\left(f_k^{-1}\left(Y_m\right) \right)=z_r \right) = \frac{1}{R}.
\end{equation}

Furthermore, let \(\hat{\mathbf{W}}=\left(\hat{W}_1, \hat{W}_2, \dots, \hat{W}_R \right)\), where \(\hat{W}_r\) denotes the number of votes mapped to \(z_r\) in the case where the total number of voters is assumed to be a Poisson random variable with parameter \(M\). By exploiting the thinning property of Poisson RVs \cite{thinning}, we have 
\begin{align}\label{eq:15111}
    \textrm{Pr}\left(\hat{\mathbf{W}}=\hat{\mathbf{w}}|X=x_k\right) &= \prod_{l=1}^{K} \textrm{Pr}\left(\hat{W_r}=\hat{w_r}|X=x_k\right)\nonumber \\
                                                                    &=\prod_{r=1}^{R} \frac{e^{-\frac{M}{R}} \left(\frac{M}{R}\right)^{\hat{w_r}}}{\hat{w_r} !},
\end{align}
which shows that \(\hat{W}_r\)'s are independent Poisson RVs. We can now use Lemma \ref{lemma:3} to show that for any event \(\mathcal{E}\) whose probability is monotonically increasing or decreasing with \(M\), we have 
\begin{equation}
    \textrm{Pr} \left(\mathcal{E}\left(\mathbf{W}\right) \right) \leq 2 \textrm{Pr} \left(\mathcal{E}\left(\hat{\mathbf{W}}\right) \right).
\end{equation}

Recalling that the number of votes submitted for \(x_l\), \(V_l\), is given by
\begin{equation}
    V_l=\sum_{r\in f_k^{-1}\left(x_1\right)}W_r,
\end{equation}
any event on \(\mathbf{V}\) (or \(\hat{\mathbf{V}}\)) is uniquely mapped to an event on \(\mathbf{W}\) (or \(\hat{\mathbf{W}}\)). More specifically, 
\begin{align}
    \textrm{Pr} \left(\mathcal{E}\left(V_1,V_2,\dots,V_K\right) \right)
                                                                    &=\textrm{Pr} \resizebox{.522 \hsize}{!}{\(\left(\mathcal{E}\left(  \sum_{r\in f_k^{-1}\left(x_1\right)}W_r, \dots, \sum_{r\in f_k^{-1}\left(x_K\right)}W_r \right)\right)\)} \nonumber \\
                                                                    &\leq 2 \textrm{Pr}\resizebox{.5 \hsize}{!}{\(\left(\mathcal{E}\left( \sum_{r\in f_k^{-1}\left(x_1\right)}\hat{W}_r, \dots, \sum_{r\in f_k^{-1}\left(x_K\right)}\hat{W}_r \right)\right) \)}\nonumber \\
                                                                    &\stackrel{(a)}{=} 2 \textrm{Pr}\left(\mathcal{E}\left(\hat{V}_1,\hat{V}_2, \dots,\hat{V}_K \right)\right),
\end{align}
where \((a)\) is obtained from the fact that the sum of independent Poisson RVs is Poisson. Therefore, the proof is complete. 

\section{Proof Of Theorem \ref{prop1}}
Using Lemma \ref{lemma:4}, we bound \(\textrm{Pr}\left(V_{l_1}\leq V_{l_2}|X=x_k\right)\) as
\begin{equation}\label{eq:17}
    \textrm{Pr}\left(V_{l_1}\leq V_{l_2}|X=x_k\right)\leq 2 \textrm{Pr}\left(\hat{V}_{l_1}\leq\hat{V}_{l_2}|X=x_k\right),
\end{equation}
where \(\hat{V}_{l_1}\) and \(\hat{V}_{l_2}\) are independent Poisson RVs with parameter \(\lambda_{l_1|k}=M p_{l_1|k}\) and \(\lambda_{l_2|k}=Mp_{l_2|k}\), respectively. Next, we use Chernoff bound to derive an upper bound on \(\textrm{Pr}\left(\hat{V}_{l_1}\leq\hat{V}_{l_2}|X=x_k\right)\), as
\begin{align}\label{eq:34}
     \textrm{Pr}\left(\hat{V}_{l_1}\leq\hat{V}_{l_2}|X=x_k\right) &\leq E\left\{e^{t\left(\hat{V}_{l_2}-\hat{V}_{l_1}\right)}\right\} \nonumber\\
                                  &= E\left\{e^{t\hat{V}_{l_2}}\right\} E\left\{e^{-t\hat{V}_{l_1}}\right\}\nonumber\\
                                  &= e^{\lambda_{l_2|k}\left(e^t-1\right) + \lambda_{l_1|k}\left(e^{-t}-1\right)}.
\end{align}
By taking the derivative of right hand side of \eqref{eq:34} and solving it for \(t\), we find \(t_0\), which is the value of \(t\) for \(t>0\) that minimizes \eqref{eq:17}, as
\begin{equation}\label{eq:50}
    t_0=\ln{\frac{\sqrt{\lambda_{l_2|k} \lambda_{l_1|k}}}{\lambda_{l_2|k}}}. 
\end{equation}
Please note that \(p_{l_1|k}>p_{l_2|k}\), hence, \(t_0=\ln{\frac{\sqrt{\lambda_{l_2|k} \lambda_{l_1|k}}}{\lambda_{l_2|k}}}\) is a positive and satisfies \(t>0\). 

By substituting \(t\) in \eqref{eq:34} with \(t_0\) in \eqref{eq:50}, we have for \eqref{eq:34} the following
\begin{align}
     \textrm{Pr}\left(V_{l_1}\leq V_{l_2}|X=x_k\right)&\leq 2 \textrm{Pr}\left(\hat{V}_{l_1}\leq\hat{V}_{l_2}|X=x_k\right) \nonumber\\ 
                                 &\leq 2 e^{-\left(\sqrt{\lambda_{{l_1}|k}}- \sqrt{\lambda_{l_2|k}}\right)^2}\nonumber\\
                                 &= 2 e^{-M\left(\sqrt{ p_{{l_1}|k}}- \sqrt{ p_{l_2|k}}\right)^2}.
\end{align}

\section{Proof Of Theorem \ref{theorem:11}}
We can bound the modified Bessel function of the first kind, \(I_{\left|\alpha\right|}\left(x\right)\), as \cite{kasperkovitz1980asymptotic}
\begin{align}\label{eq:80}
    I_{\left|\alpha\right|}\left(x\right) = \frac{e^x e^{\frac{-\alpha^2}{x}}}{\sqrt{2\pi x}}\left(1+\epsilon_0\right)\leq \frac{e^x}{\sqrt{2\pi x}}\left(1+\epsilon_0\right),
\end{align}
where \(|\epsilon_0|\) is of order \(x^{\frac{-1}{4}}\) \cite{kasperkovitz1980asymptotic}. Therefore, we have 
\begin{equation}
    I_{\left|\alpha\right|}\left(2M\sqrt{p_{l|k}p_{k|k}}\right) \leq  \frac{e^{2M\sqrt{p_{l|k}p_{k|k}}} \left(1+\epsilon_0\right)}{\sqrt{2\pi2M\sqrt{p_{l|k}p_{k|k}}}}.
\end{equation}
By using \eqref{eq:80} to substitute \(I_{\left|\alpha\right|}\left(2M\sqrt{p_{l|k}p_{k|k}}\right)\) in \eqref{eq:upper-bound}, we obtain the inequality in \eqref{eq:61}, 
\begin{figure*}
  \hrulefill
\begin{align}\label{eq:61}
    P_e &\leq 2 \left(1-\frac{1}{K} \sum _{k=1} ^K \prod_{\substack{l=1 \\ l \neq k}}^K \left(1-\sum_{\alpha=0}^\infty e^{-M\left(p_{k|k}+p_{l|k}\right)}\left(\frac{p_{l|k}}{p_{k|k}}\right)^{\frac{\alpha}{2}}\frac{e^{\left(2M\sqrt{p_{l|k}p_{k|k}}\right)}\left(1+\epsilon_0\right)}{\sqrt{2\pi2M\sqrt{p_{l|k}p_{k|k}}}}\right)\right) \nonumber\\
        &= 2 \left(1-\frac{1}{K} \sum _{k=1} ^K \prod_{\substack{l=1 \\ l \neq k}}^K \left(1- \frac{e^{-M\left(\sqrt{p_{k|k}}-\sqrt{p_{l|k}}\right)^2}\left(1+\epsilon_0\right)}{\sqrt{2\pi2M\sqrt{p_{l|k}p_{k|k}}}}   \sum_{\alpha=0}^\infty\left(\frac{p_{l|k}}{p_{k|k}}\right)^{\frac{\alpha}{2}}\right)\right) \nonumber\\
        &\stackrel{(a)}{=} 2\left(1-\frac{1}{K} \sum _{k=1} ^K \prod_{\substack{l=1 \\ l \neq k}}^K \left(1- \frac{e^{-M\left(\sqrt{p_{k|k}}-\sqrt{p_{l|k}}\right)^2}\left(1+\epsilon_0\right)}{\sqrt{2\pi2M\sqrt{p_{l|k}p_{k|k}}}\left( 1-\sqrt{\frac{p_{l|k}}{p_{k|k}}}\right)}  \right)\right).
\end{align}
  \hrulefill
\end{figure*}
where \((a)\) in \eqref{eq:61} follows from the following equality
\begin{equation}
    \sum_{\alpha=0}^\infty\left(\frac{p_{l|k}}{p_{k|k}}\right)^{\frac{\alpha}{2}}=\frac{1}{1-\sqrt{\frac{p_{l|k}}{p_{k|k}}}}\nonumber.
\end{equation} 
We upper bound the right-hand-side of \eqref{eq:61} to obtain
\begin{align}\label{eq:63}
    P_e &\stackrel{(b)}{\leq} \resizebox{.9 \hsize}{!}{\( 2 \left(1- \prod_{\substack{l=1 \\ l \neq k^*}}^K   \left(1- \frac{e^{-M\left(\sqrt{p_{k^*|k^*}}-\sqrt{p_{l|k^*}}\right)^2}\left(1+\epsilon_0\right)}{\sqrt{2\pi2M\sqrt{p_{l|k^*}p_{k^*|k^*}}}\left( 1-\sqrt{\frac{p_{l|k^*}}{p_{k^*|k^*}}}\right)} \right)\right)\)}\nonumber \\
        &\stackrel{(c)}{\leq} 2 \left(\sum_{\substack{l=1 \\ l \neq k^*}}^K \frac{e^{-M\left(\sqrt{p_{k^*|k^*}}-\sqrt{p_{l|k^*}}\right)^2}\left(1+\epsilon_0\right)}{\sqrt{4\pi M\sqrt{p_{l|k^*}p_{k^*|k^*}}}\left( 1-\sqrt{\frac{p_{l|k^*}}{p_{k^*|k^*}}}\right)} \right),
\end{align}
where \((b)\) follows from the fact that substituting \(k\) with \(k^*\) in \eqref{eq:61} maximizes the right-hand-side of the inequality and \((c)\) is obtained using the following inequality. Let \(a_1,a_2,\dots, a_N\) be real numbers in \([0,1]\), then we have 
\begin{equation} \label{eq:1021}
     1-\prod_{n=1}^N \left(1-a_n\right) \leq \sum_{n=1}^N a_n \nonumber.
\end{equation}
This inequality is known as Weierstrass Inequality and can be proved using induction~\cite{bromwich2005introduction}. 

Next, we upper bound the right-hand-side of \eqref{eq:63} to obtain 
\begin{equation}\label{eq:201}
    P_e\leq 2 \left(K-1\right) \frac{e^{-M\left(\sqrt{p_{k^*|k^*}}-\sqrt{p_{l^*|k^*}}\right)^2}\left(1+\epsilon_0\right)}{\sqrt{4\pi M\sqrt{p_{l^*|k^*}p_{k^*|k^*}}}\left( 1-\sqrt{\frac{p_{l^*|k^*}}{p_{k^*|k^*}}}\right)},
\end{equation}
where this upper bound follows from the fact that substituting \(l\) by \(l^*\) in the typical element of the summation maximizes the right-hand-side of \eqref{eq:63}. 
Finally, we take the logarithm of both sides and divide the result by \(M\) to obtain the slope of the decay of error rate versus the number of voters as 
\begin{equation}\label{eq:65}
    \resizebox{.992 \hsize}{!}{\(\lim_{M\to \infty}\frac{\ln{P_e}}{M} \leq \lim _{M\to \infty}\frac{1}{M}\left(\ln{\gamma}-M\left(\sqrt{p_{k^*|k^*}}-\sqrt{p_{l^*|k^*}} \right)^2 - \frac{1}{2}\ln{M} \right)\)},
\end{equation}
where \(\gamma=\frac{2\left(K-1\right)\left(1+\epsilon_0\right)}{\sqrt{4\pi \sqrt{p_{l|k^*}p_{k^*|k^*}}}\left( 1-\sqrt{\frac{p_{l|k^*}}{p_{k^*|k^*}}}\right)}\). Therefore, 
\begin{equation}\label{eq:66}
    \lim_{M\to \infty}\frac{\ln{P_e}}{M} \leq -\left(\sqrt{p_{k^*|k^*}}-\sqrt{p_{l^*|k^*}} \right)^2 +  \epsilon, 
\end{equation}
where \(\epsilon = \lim_{M\to \infty} \frac{1}{M} \left(\ln{\gamma} - \frac{1}{2}\ln{M} \right)=0\), since \(O\left(\epsilon\right)=\frac{O\left(\epsilon_0 \right)-\ln\left(M^{-1/2}\right)}{M}=\frac{\ln\left(M^{-3/4}\right)}{M}\). Hence, the proof is complete.

\begin{figure*}[t]
  \hrulefill
\begin{align}\label{eq:4001}
    \lim_{M\to \infty} \textrm{Pr}\left(V_{l_1}<V_{l_2}|X=x_k\right)&\stackrel{(a)}{\leq} \lim_{M\to\infty}\textrm{Pr}\left(\bigcup_{t=1}^T V_{l_2}^{(t)}-V_{l_1}^{(t)} \geq E\{V_{l_2}^{(t)}-V_{l_1}^{(t)}\} + M r_t \delta _{l_1,l_2|k}|X=x_k\right)\nonumber\\
    &\stackrel{(b)}{\leq} \lim_{M\to \infty} \sum_{t=1}^T \textrm{Pr}\left(V_{l_2}^{(t)}-V_{l_1}^{(t)} \geq E\{V_{l_2}^{(t)}-V_{l_1}^{(t)}\} + M r_t \delta_{l_1,l_2|k}|X=x_k\right)\nonumber\\
    &\stackrel{(c)}{\leq} \lim_{M\to \infty} \sum_{t=1}^T \biggl(\textrm{Pr}\left(V_{l_2}^{(t)}-V_{l_1}^{(t)} - E\{V_{l_2}^{(t)}-V_{l_1}^{(t)}\} \geq M r_t \delta _{l_1,l_2|k}|X=x_k\right)\nonumber\\
    &\textrm{~~~~~~~~~~~~~~~~}+\textrm{Pr}\left( V_{l_2}^{(t)}-V_{l_1}^{(t)} - E\{V_{l_2}^{(t)}-V_{l_1}^{(t)}\} \leq -M r_t \delta _{l_1,l_2|k}|X=x_k\right) \biggl) \nonumber\\
    &\stackrel{(d)}{=} \sum_{t=1}^T \lim_{M\to \infty}\textrm{Pr}\left(|V_{l_2}^{(t)}-V_{l_1}^{(t)} - E\{V_{l_2}^{(t)}-V_{l_1}^{(t)}\}| \geq M r_t \delta _{l_1,l_2|k}|X=x_k\right)\nonumber\\
    &\stackrel{(e)}{=} 0, 
\end{align}
\hrulefill
    \begin{align}\label{eq:4002}
        Pr \left(\hat{X}\neq X\right)&\stackrel{(a)}{=} \frac{1}{K}\sum_{k=1}^K \left(1-\prod_{l=1}^K \left(1-\textrm{Pr}\left(\hat{X}=x_l|X=x_k\right)\right)\right)\nonumber\\
        &\stackrel{(b)}{=} \frac{1}{K}\sum_{k=1}^K \left(1- \prod_{l=1}^K \prod_{t=1}^T  \left(1-\textrm{Pr}\left({V}_l^{(t)}\geq E\{{V}_l^{(t)}\}+\frac{r_t M \delta_{l|k}}{2} \right)\right) \left(1-\textrm{Pr}\left({V}_k^{(t)}\leq E\{{V}_k^{(t)}\}-\frac{r_t M \delta_{l|k}}{2} \right)\right)\right)\nonumber\\
        &\stackrel{(c)}{\leq} \frac{1}{K}\sum_{k=1}^K \left(1- \prod_{l=1}^K \prod_{t=1}^T 
                \left(1- \exp{\left(\frac{-\left(r_t M\right) ^2\delta_{l|k}^2}{8\left(E\{{V}_l^{(t)}\}+\frac{r_t M \delta_{l|k}}{6} \right)} \right)} \right) \left(1- \exp{\left(\frac{-\left(r_t M\right) ^2\delta_{l|k}^2}{8E\{V_k^{(t)}\} } \right)}\right) \right)\nonumber\\
        &\stackrel{(d)}{=} \frac{1}{K}\sum_{k=1}^K \left(1- \prod_{l=1}^K \prod_{t=1}^T 
                        \left(1- \exp{\left(\frac{-r_t M \delta_{l|k}^2}{8\left(q_{l|K}^{(t)}+\frac{\delta_{l|k}}{6} \right)} \right)} \right) \left(1- \exp{\left(\frac{-r_t M \delta_{l|k}^2}{8q_{k|k}^{(t)} } \right)}\right)
                                       \right)\nonumber\\
        &\stackrel{(e)}{\leq} \frac{1}{K}\sum_{k=1}^K \left(\sum_{l=1}^K \sum_{t=1}^T \left(\exp{\left(\frac{-r_t M \delta_{l|k}^2}{8\left(q_{l|K}^{(t)}+\frac{\delta_{l|k}}{6} \right)} \right)}+ \exp{\left(\frac{-r_t M \delta_{l|k}^2}{8q_{k|k}^{(t)} } \right)}\right) \right)\nonumber \\
        &\stackrel{(f)}{\leq} 2KT \left(\exp{\left(-M\epsilon^*\right)} \right),
    \end{align}
      \hrulefill
\end{figure*}

\section{Proof of Lemma \ref{lemma:5}} 
We prove Lemma \ref{lemma:5} through contradiction. To this end, without loss of generality, let's assume that the MVF outputs class \(x_{l^*}\) where \(x_{l^*} \in \mathcal{X}\), \(x_{l^*}\neq x_k\), and concurrently we have 
\begin{align} \label{eq:305}
   {v}_l^{(t)}-{v}_k^{(t)} < E\{{V}_l^{(t)}-{V}_k^{(t)}|&X=x_k\}+r_t M \delta_{l|k},\nonumber\\
   & \forall\, l\in [K] \text{~and~} \forall \, t \in [T].
\end{align}
Since \eqref{eq:305} holds \(\forall l \in [K]\) and \(\forall t \in [T]\), we can take the summation of both sides of \eqref{eq:305} with respect to \(t\) and obtain
\begin{align}\label{eq:810}
    \sum_{t=1}^T \left({v}_l^{(t)}-{v}_k^{(t)}\right) &< \sum_{t=1}^T \resizebox{.61 \hsize}{!}{\(\left(E\{{V}_l^{(t)}-{V}_k^{(t)}|X=x_k\}+r_t M \delta_{l|k} \right)\)} \nonumber\\
                                                      & = \sum_{t=1}^T \resizebox{.61 \hsize}{!}{\(E\{{V}_l^{(t)}-{V}_k^{(t)}|X=x_k\}+ \sum_{t=1}^T r_t M \delta_{l|k}\)}  \nonumber\\
                                                      & \stackrel{(a)}{=} \sum_{t=1}^T \resizebox{.6 \hsize}{!}{\(\left(r_t M  q_{l|k}^{(t)} - r_t M q_{k|k}^{(t)}\right) + M \delta_{l|k}\sum_{t=1}^T r_t \)} \nonumber\\
                                                      & = M \sum_{t=1}^T r_t  \left(q_{l|k}^{(t)} - q_{k|k}^{(t)}\right) + M \delta_{l|k} \nonumber\\
                                                      &\stackrel{(b)} = 0,
\end{align}
where \((a)\) is obtained using the fact that \(\mathbf{V}^{(t)}\) is a Multinomial random vector, and as a result, \(E\{{V}_l^{(t)}|X=x_k\}= r_t M q_{l|k}^{(t)}\), \(\forall l,k \in [K]\). Next, \((b)\) is obtained by substituting \(\delta_{l|k}\) from \eqref{eq:301}.
Therefore, \eqref{eq:810} shows that \(\forall l\in[K]\), including \(l^*\), we have
\begin{equation}
    v_l - v_k = \sum_{t=1}^T \left(V_l^{(t)}-V_k^{(t)}\right) < 0, \text{~} \forall l \in [K], 
\end{equation} 
which contradicts the assumption that the MVF incorrectly outputs class \(x_{l^*}\). Therefore, the inequality in \eqref{eq:305} cannot be satisfied and the proof is complete.

\section{Proof Of Theorem \ref{prop:10}}
To prove Theorem \ref{prop:10}, we bound \(\lim_{M\to \infty}\textrm{Pr}\left(V_{l_1}<V_{l_2}|X=x_k\right) \) as presented in \eqref{eq:4001}, 
where \((a)\) in \eqref{eq:4001} is a result of Lemma \ref{lemma:5}, \((b)\) is obtained using the union bound technique, \((c)\) is derived by adding a positive value to the typical element of the summation in \((b)\), and \((d)\) is resulted from combining the two probability expressions in \((c)\) into one probability expression. Finally, \((e)\) is due to the weak law of large numbers \cite{cover} used to bound the difference of RV \(V_{l_1}-V_{l_2}\) and its mean value.

\section{Proof Of Theorem \ref{prop:3}}
We bound the error rate as presented in \eqref{eq:4002}, where \((a)\) in \eqref{eq:4002} is obtained from the definition of the error rate. 
Moreover, \((b)\) follows from Lemma \ref{lemma:5} which states that if \(\hat{X}\neq X\) while \(X= x_k\), there exists at least one \(l \neq k \in [K]\) such that 
    \begin{equation}\label{eq:1020}
         {V}_l^{(t)}-{V}_k^{(t)} \geq E\{{V}_l^{(t)}-{V}_k^{(t)}\}+r_t M \delta_{l|k}.
    \end{equation}
If inequality \eqref{eq:1020} holds, at least one of the following inequalities must hold
    \begin{equation}\label{eq:803}
        \begin{cases}
            &{V}_l^{(t)}\geq E\{{V}_l^{(t)}\}+\frac{r_t M \delta_{l|k}}{2},\\
            &{V}_k^{(t)}\leq E\{{V}_k^{(t)}\}-\frac{r_t M \delta_{l|k}}{2}.
         \end{cases}
    \end{equation}
Therefore, \((b)\) is a result of \eqref{eq:803}. Additionally, \((c)\) is obtained using Chernoff bound on the upper and lower tail of Binomial RVs \cite{Chernoff}, \(V_{l|k}^{(t)}\) and \(V_{k|k}^{(t)}\), \((d)\) is resulted from substituting \(E\{V_l^{(t)}\}\) and \(E\{V_k^{(t)}\}\) by \(r_t M q_{l|k}^{(t)}\) and \(r_t M q_{k|k}^{(t)}\), respectively. Furthermore, \((e)\) is a result of Weierstrass Inequality induction~\cite{bromwich2005introduction} presented in \eqref{eq:1021}, and \((f)\) is obtained using
\begin{align}
    \exp{\left(\frac{-r_t M \delta_{l|k}^2}{8\left(q_{l|K}^{(t)}+\frac{\delta_{l|k}}{6} \right)} \right)}+ &\exp{\left(\frac{-r_t M \delta_{l|k}^2}{8q_{k|k}^{(t)} } \right)} \nonumber\\ \leq 2 \exp\left(-M\epsilon^*\right),\,
    &\textrm{~}\forall \, l,k \in [K], \text{~and~} \forall \, t \in [T]. 
\end{align}

\end{document}